\documentclass{article} % For LaTeX2e
\usepackage{iclr2025_conference,times}

\usepackage{amsmath,amsfonts,bm}

\def\eqref#1{equation~\ref{#1}}
\def\1{\bm{1}}

\DeclareMathAlphabet{\mathsfit}{\encodingdefault}{\sfdefault}{m}{sl}
\SetMathAlphabet{\mathsfit}{bold}{\encodingdefault}{\sfdefault}{bx}{n}

\usepackage{hyperref}
\usepackage{url}
\usepackage{paralist}
\usepackage{graphicx}
\usepackage{booktabs}
\usepackage{wrapfig}
\usepackage{subcaption}
\usepackage{supertabular}
\usepackage{pgfplots}
\usepackage{multirow}
\usepackage{tcolorbox}

\usepackage{CJK}
\usepackage{algorithm}
\usepackage{algorithmicx}
\usepackage{algpseudocode}
\usepackage{amsmath}
\usepackage{pgfplots}

\pgfplotsset{compat=1.18}
\floatname{algorithm}{algorithm}

\usepackage[capitalize,noabbrev]{cleveref}

\usepackage{listings}

\definecolor{codegreen}{rgb}{0,0.6,0}
\definecolor{codegray}{rgb}{0.5,0.5,0.5}
\definecolor{codepink}{RGB}{252, 142, 172}
\definecolor{codepurple}{rgb}{0.58,0,0.82}
\definecolor{backcolour}{RGB}{245,245,245}
\lstdefinestyle{mystyle}{
    backgroundcolor=\color{backcolour},   
    commentstyle=\color{magenta},
    keywordstyle=\color{blue},
    numberstyle=\tiny\color{codegray},
    stringstyle=\color{codepurple},
    basicstyle=\fontfamily{\ttdefault}\footnotesize,
    breakatwhitespace=false,        
    breaklines=true,                
    keepspaces=true,    
    frame=single,
    numbersep=5pt,                  
    showspaces=false,              
    showstringspaces=false,
    showtabs=false,               
    tabsize=2,
    classoffset=1, % starting new class
    keywordstyle=\color{violet},
    classoffset=0,
}
\hypersetup{
	colorlinks=true,%
	citecolor={blue!50!black},
	linkcolor={red!50!red},
	urlcolor={green!50!black}
}

\newcommand{\ie}{\emph{i.e., }}
\newcommand{\eg}{\emph{e.g., }}

\newcommand{\projname}{\textbf{MART}}
\newcommand{\projnameALL}{\textbf{M}LLM \textbf{A}s \textbf{R}e\textbf{T}riever}
\newcommand{\aithor}{AI2-THOR}
\newcommand{\legent}{LEGENT}
\newcommand{\PlainAgent}{\textbf{PA}}
\newcommand{\SimilarityLLaVA}{\textbf{SL}}

\title{MLLM as Retriever: Interactively Learning Multimodal Retrieval for Embodied Agents}

\author{%
  Junpeng Yue$^1$\thanks{Junpeng Yue and Xinrun Xu work as interns at BAAI.}, \ Xinrun Xu$^{2}$, Börje F. Karlsson$^3$, and Zongqing Lu$^{1}$\thanks{Correspondence to Zongqing Lu \textless zongqing.lu@pku.edu.cn\textgreater.} \\  
   $^1$School of Computer Science, Peking University \\
   $^2$Institute of Software, Chinese Academy of Sciences \\
   $^3$Beijing Academy of Artificial Intelligence
}

\iclrfinalcopy
\usepackage{hyperref}
\begin{document}

\maketitle

\begin{abstract}

MLLM agents demonstrate potential for complex embodied tasks by retrieving multimodal task-relevant trajectory data. 
However, current retrieval methods primarily focus on surface-level similarities of textual or visual cues in trajectories, neglecting their effectiveness for the specific task at hand.
To address this issue, we propose a novel method, \projnameALL\ (\projname), which enhances the performance of embodied agents by utilizing interaction data to fine-tune an MLLM retriever based on preference learning, such that the retriever fully considers the effectiveness of trajectories and prioritize them for unseen tasks.
We also introduce Trajectory Abstraction, a mechanism that leverages MLLMs' summarization capabilities to represent trajectories with fewer tokens while preserving key information, enabling agents to better comprehend milestones in the trajectory.
Experimental results across various environments demonstrate our method significantly improves task success rates in unseen scenes compared to baseline methods.
This work presents a new paradigm for multimodal retrieval in embodied agents, by fine-tuning a general-purpose MLLM as the retriever to assess trajectory effectiveness. All the code for benchmark tasks, simulator modifications and the MLLM retriever is available at \href{https://github.com/PKU-RL/MART}{https://github.com/PKU-RL/MART}.
\end{abstract}

% \vspace{5 ex}

\begin{figure}[ht]
    \centering
    % \vspace{-4mm}
    \includegraphics[width=.95\linewidth]{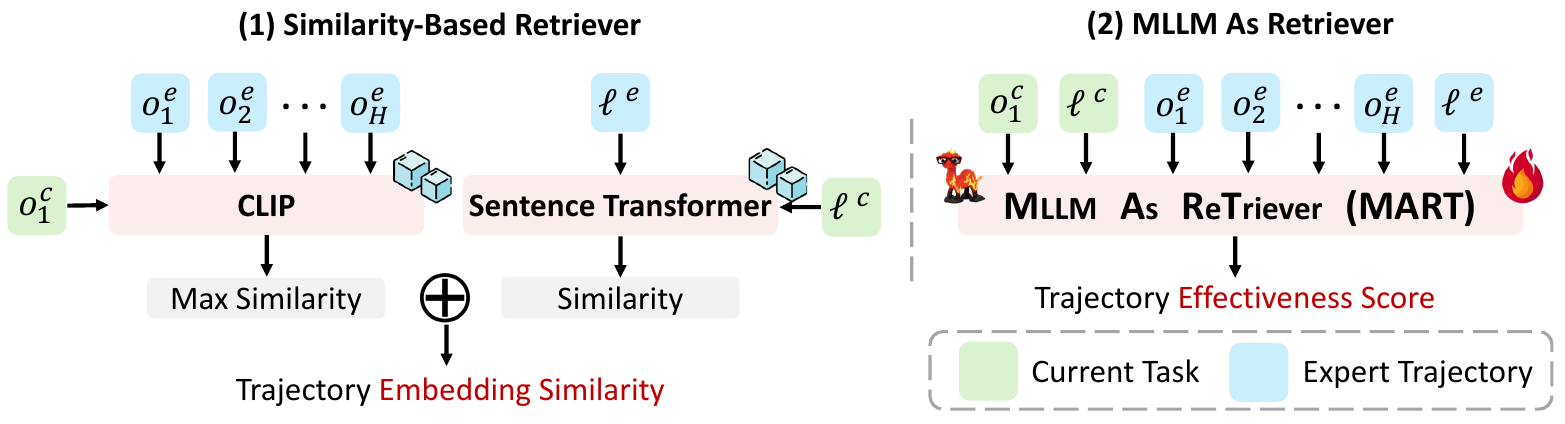}
    \caption{Similarity-Based Retriever vs. \projname. Traditional multimodal retrieval methods (1) depend on calculating weighted sums of image and text embedding similarities, while our approach (2) introduces interactive learning to assess the relevance between the current and expert trajectories.}
    \label{fig:comparison_simple}
    \vspace{-4pt}
\end{figure}

\section{Introduction}
\label{sec:introduction}

Embodied agents interacting with complex environments require understanding both the current context and task-specific domain knowledge to perform effectively~\citep{wang_jarvis-1_2023s,lifshitz2023steve}. Recently, Multimodal Large Language Models (MLLMs), which are capable of processing both textual and visual data, have shown promise in various embodied tasks -- \eg table manipulation~\citep{handa2023dextreme,chen2022system}, robot navigation~\citep{zhang2024navidvideobasedvlmplans,LM-Nav}, and 3D games~\citep{wangvoyager,tan2024cradle,jiang2024reinforcement}. 
However, such models typically lack effective grounding in the embodied environments in which agents operate, greatly limiting their performance in embodied tasks \citep{DiscussNav,wang2024llm3}.

% To mitigate this limitation, providing additional task-relevant domain knowledge is essential to better leverage MLLMs capabilities. Trajectory data, consisting of sequences of actions and observations, can be easily available and provide valuable insights into task execution. By using trajectory data as a component in prompting an MLLM, one can readily leverage previously encountered experiences to better guide agents through similar tasks in new situations or environments. However, selecting the most effective trajectories — those that can significantly enhance task performance — remains a challenge, particularly when multiple trajectories appear similar in both textual and visual modalities.

To mitigate this limitation, providing additional task-relevant grounding information is essential to better leverage the general capabilities of MLLMs. Trajectory data, consisting of sequences of actions and observations, can be easily available and provide valuable insights into task execution \citep{zheng2023synapse,zhao2024expel}, therefore serving as a good information source for grounding. By using trajectory data in prompting an MLLM, the embodied agent can readily leverage previous experiences to better guide agents through similar tasks in new situations or environments \citep{zhang2024mobileexperts,lee2024exploreselectderiverecall}. However, retrieving the most effective trajectories — those that can significantly enhance task performance — remains a challenge, particularly when multiple trajectories appear similar in both textual and visual modalities \citep{jeurissen2024playingnethack}.

\begin{figure}[ht]
\centering
\begin{minipage}{.95\textwidth}
        \centering
        \subcaptionbox{\aithor}[0.47\linewidth]{
            \includegraphics[width=0.49\linewidth]{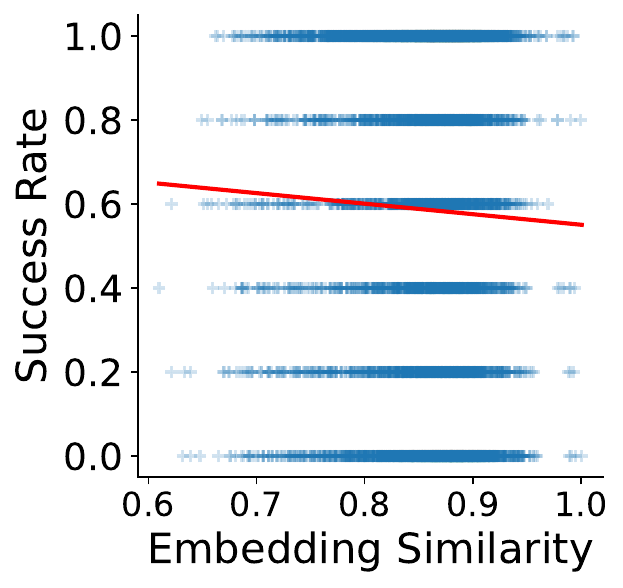}
            \includegraphics[width=0.48\linewidth]{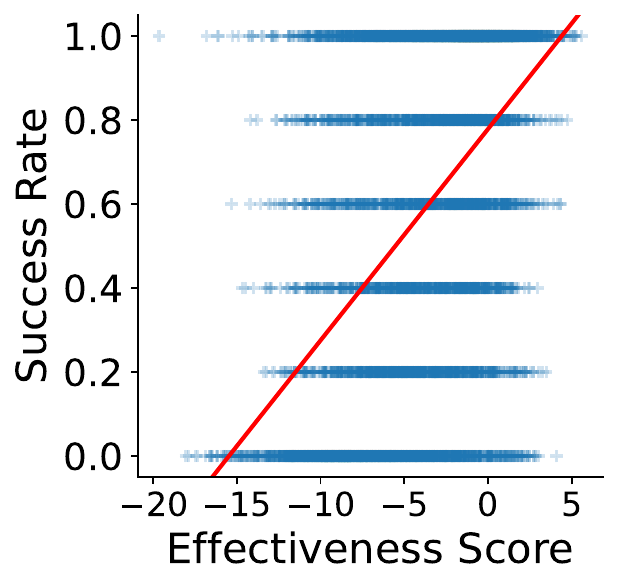}
        }
        \hspace{5mm}
        \subcaptionbox{\legent}[0.47\linewidth]{
            \includegraphics[width=0.49\linewidth]{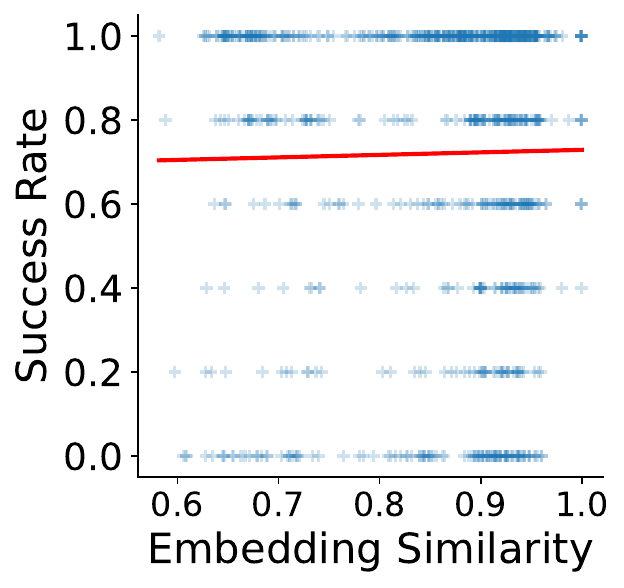}
            \includegraphics[width=0.49\linewidth]{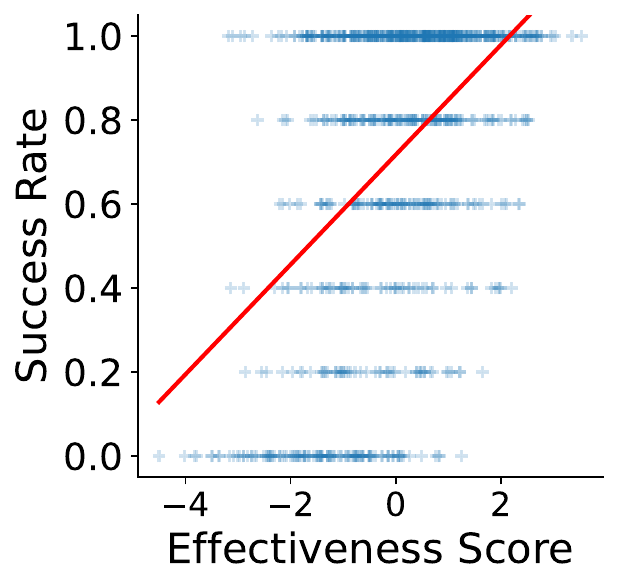}
        }
    \end{minipage}
    \vspace{-2mm}
\caption{Scatter plots illustrating the relationship between success rate and embedding similarity (left) or effectiveness score (right) in two environments. The red line indicates a linear fit to the data.}
\label{fig:scatter}
\end{figure}

Existing retrieval methods mainly focus on surface-level textual or visual similarities of trajectories, often neglecting key aspects critical for task effectiveness, \eg a trajectory with a similar task instruction but in a different scene, or one in the same scene but with a different layout. In such cases, these trajectories fail to provide %genuinely 
useful information for the current task and can mislead the agent. 
As shown in Figure \ref{fig:scatter}, relying solely on similarity is not effective in retrieving useful trajectories, as similarity does not directly correlate with success rate.
% To better support agents in embodied tasks, retrievers need to retrieve trajectories that are more contextually grounded in the environment, requiring the incorporation of interactive learning.
To better support agents in embodied tasks, a trajectory retriever model needs to consider the effectiveness of trajectories for a given task. %, necessitating the incorporation of interactive learning.

To achieve a better retriever, we propose a new paradigm that integrates interactive learning with the retriever. Firstly, we consider expert trajectories of training scenarios as prompt for an MLLM agent, and let the agent interact with the environment to collect different success rates for different such reference trajectories. This interactive feedback data is then organized into preference pairs, which are used to fine-tune an MLLM -- LLaVA~\citep{NEURIPS2023_6dcf277e} in our case -- with a Bradley-Terry model~\citep{bradley1952rank}, such that the fine-tuned retriever model is capable of prioritizing more effective trajectories for unseen tasks. Combining this functionality with the inherent general capabilities of MLLM allows embodied agents to operate more effectively in unseen environments by leveraging their most useful past experiences.

We also introduce a new Trajectory Abstraction mechanism, which uses MLLMs' summarization capabilities to represent trajectories in a reduced number of tokens, while preserving key information and enabling agents to better understand such information in the trajectory (\eg key relevant overarching actions). This mechanism is especially important in long-horizon tasks, both reducing the required context window length and removing distracting information from trajectory samples.

Combining the aforementioned components, we present our approach -- \projname\ (\projnameALL) -- which adapts embodied agents in unseen scenarios by fine-tuning MLLM through preference data.
To assess the benefits of our method, we conduct empirical experiments across diverse environments. The experimental results show that \projname\ achieves significantly higher task success rates compared to baselines, demonstrating its effectiveness. With this approach, we present a new paradigm for multimodal retrieval in embodied agents, fine-tuning a general-purpose MLLM as a retriever capable of considering trajectory effectiveness. Our contributions can be summarized as follows:  
% in different environments that demonstrate our  where the retrieval-augmented MLLM agent achieves significantly higher task success rates compared to baseline methods. Our contributions include:
\begin{itemize}
    \item To the best of our knowledge, \projname\ is the first approach that integrates interactive learning with a retriever and uses interactive feedback to fine-tune an MLLM retriever in evaluating trajectory effectiveness, combining its inherent general capabilities with the ability to assess the task-guiding effectiveness of trajectories.
    % \item To our knowledge, \projname\ is the first approach that integrates interactive learning with a retriever and uses interactive feedback to fine-tune MLLM retriever in evaluating trajectory effectiveness, utilizing a reward model from RLHF as the retriever;
    \item We introduce Trajectory Abstraction, a new mechanism that utilizes MLLM capabilities to significantly condense trajectories. This method reduces the token number while retaining essential information, allowing agents to effectively use this condensed knowledge in novel situations and provide guidance for long-horizon tasks.
    % \item \projname\ is empirically verified by comprehensive experiments across environments, which showcase the effectiveness and significant performance benefits of the proposed framework on unseen tasks. \projname greatly surpass baselines over 10 percentages across environments.
    \item The effectiveness of \projname\ is empirically validated through comprehensive experiments in various environments, demonstrating significant performance improvements on unseen tasks. \projname\ consistently surpasses baselines by over 10\% across different environments.
\end{itemize}

% \borje{add numerical results to intro/contributions} \junpeng{Done. And I modified the motivation of intro}

% 1. Innovatively propose to use interactive feedbacks to train MLLM, so that MLLM can be used to evaluate the effectiveness of trajectories and use the reward model of RLHF as retriever.

% 2. Propose a new way to store and inject context windows into trajectories, and use the automatic abstraction of MLLM to reduce the number of tokens in trajectories while maintaining key information.

% 3. Empirical validation showing improved task performance for embodied agents in unseen environments.

% With this approach, we present a new paradigm for multimodal retrieval in embodied agents, fine-tuning a general-purpose MLLM as a retriever capable of considering trajectory effectiveness.

% With this approach, we aim to bridge the gap between general-purpose MLLMs and the specific, environment- and context-driven requirements of embodied agents, offering a new paradigm for multimodal trajectory retrieval.% in the field.

\section{Related Work}
\label{sec:related_work}

% \subsection{Multimodal Embodied Agents}
\subsection{Embodied Agents Based on Large Models}
% Recently, several attempts have been made to develop multimodal embodied agents for complex tasks. 
Recently, there have been several attempts to utilize the general-purpose capabilities of large models for complex embodied tasks.
These efforts can be broadly categorized into two types: VLA models and LLM/MLLM-based agents.
\textbf{1) VLA models}, including PaLM-E~\citep{driess2023palm}, RT-2~\citep{brohan2023rt}, Gato~\citep{reed2022generalist}, VIMA~\citep{jiang2022vima}, and MOO~\citep{stone2023open}, rely on trajectory data to train a Transformer-based VLM for action planning, without explicitly constructing a memory.
However, their generalization capabilities are limited due to the inherent issue of catastrophic forgetting in neural networks.
% \textbf{1) VLA models}, such as PaLM-E~\citep{driess2023palm}, RT-2~\citep{brohan2023rt}, Gato~\citep{reed2022generalist}, VIMA~\citep{jiang2022vima}, and MOO~\citep{stone2023open}, require trajectory data to train a Transformer-based VLM for action planning, without explicitly constructing a memory. However, due to the inherent phenomenon of catastrophic forgetting in neural networks, their generalization capabilities are limited.
\textbf{2) LLM/MLLM-based agents}, like Voyager~\citep{wangvoyager} and DEPS~\citep{wang2023describe} for Minecraft, Cradle~\citep{tan2024cradle} for RDR2, LLM-Planner~\citep{song2023llmplanner} for ALFRed, and Code-as-Policies~\citep{liang2023code} for real-world embodied control, do not involve directly training new models. Instead, they leverage the general-purpose capabilities of LLM/MLLM primarily through prompt engineering.
% \textbf{2) MLLM-based agents}  include environments such as Minecraft~\citep{wang2023describe, wang_jarvis-1_2023s, wangvoyager}, Starcraft II~\citep{ma2023large}, and RDR2~\citep{tan2024cradle}. These methods do not involve training models, but instead leverage MLLM capabilities through prompt engineering. 
% They utilize visual and textual observations from internal APIs as input to MLLM, enabling actions that complete embodied tasks. 
Most of these agents build and maintain comprehensive memory systems to assist in task completion. 
However, memory retrieval mostly focuses on surface-level similarity, overlooking actual effectiveness in completing complex tasks.

\subsection{Memory Retrieval in Agents}

% Recent works have leveraged the capabilities of advanced Large Models (\eg LLM, MLLM, VLA) to function as embodied agents, such as 
% % \borje{here is should be paper, then what env they target, not use use the env name as the embodied agent reference, and add some robotis example, at least one} 
% SOFT-SC~\citep{wang-etal-2024-soft} in ALFWorld~\citep{ALFWorld20}, TWOSOME~\citep{tantrue} in VirtualHome~\citep{puig2018virtualhome}, Voyager~\citep{wangvoyager}, DEPS \citep{wang2023describe},  Jarvis-1 \citep{wang_jarvis-1_2023s} in Minecraft, \citep{ma2023large} in Starcraft II, and CoELA \citep{zhang2024building}
% in SAPIEN \citep{Xiang_2020_SAPIEN}
% with textual observations obtained from internal APIs and pre-defined semantic actions. 
Agents can continuously learn and improve by recalling task-related experiences \citep{zhang2024survey,xi2023rise}. 
During interactions with the environment, two main types of information are stored in memory. 
\textbf{1) Semantic information:} 
Early LLM agents faced limitations due to input token constraints, leading to reliance on short-term memory and greedy strategies \citep{chen2024s,zhang2024proagent,abdelnabi2023llmdeliberation,wang2023humanoid}. However, summarizing memory over short periods risked information loss \citep{light2023avalonbench,kaiya2023lyfe}. 
While storing comprehensive memory (semantic, episodic, procedural) can provide great value, its effective utilization for decision-making remains challenging \citep{Theodore2024Cognitive}. 
More recent approaches try to prioritize relevant memories based on embedding similarity or LLM-based relevance identification (\eg \citep{park2023generative}, \citep{hong2023metagpt}, \citep{AgentSims}, \citep{wangRoleLLMBenchmarkingEliciting2023}, and~\citep{xu2023exploring}).
\textbf{2) Image information:} Recent advances in MLLM agents have greatly enhanced their grounding abilities by allowing image memory retrieval during actions. As illustrated in Figure \ref{fig:comparison_simple}, Similarity-Based Retrievers for agents, \eg \citep{zhou2024minedreamer,wang_jarvis-1_2023s,li2024optimus}, leverage image embedding. %similarity in MineCLIP \citep{fan2022minedojo}. 
while Groot~\citep{cai2023groot} encodes visual and temporal information from video frames for guiding actions. But neither considers direct task effectiveness.

\section{Interactively Learning Multimodal Retrieval}
\label{sec:method}

\begin{figure}[ht]
    \centering
    % \vspace{-4mm}
    \includegraphics[width=.95\linewidth]{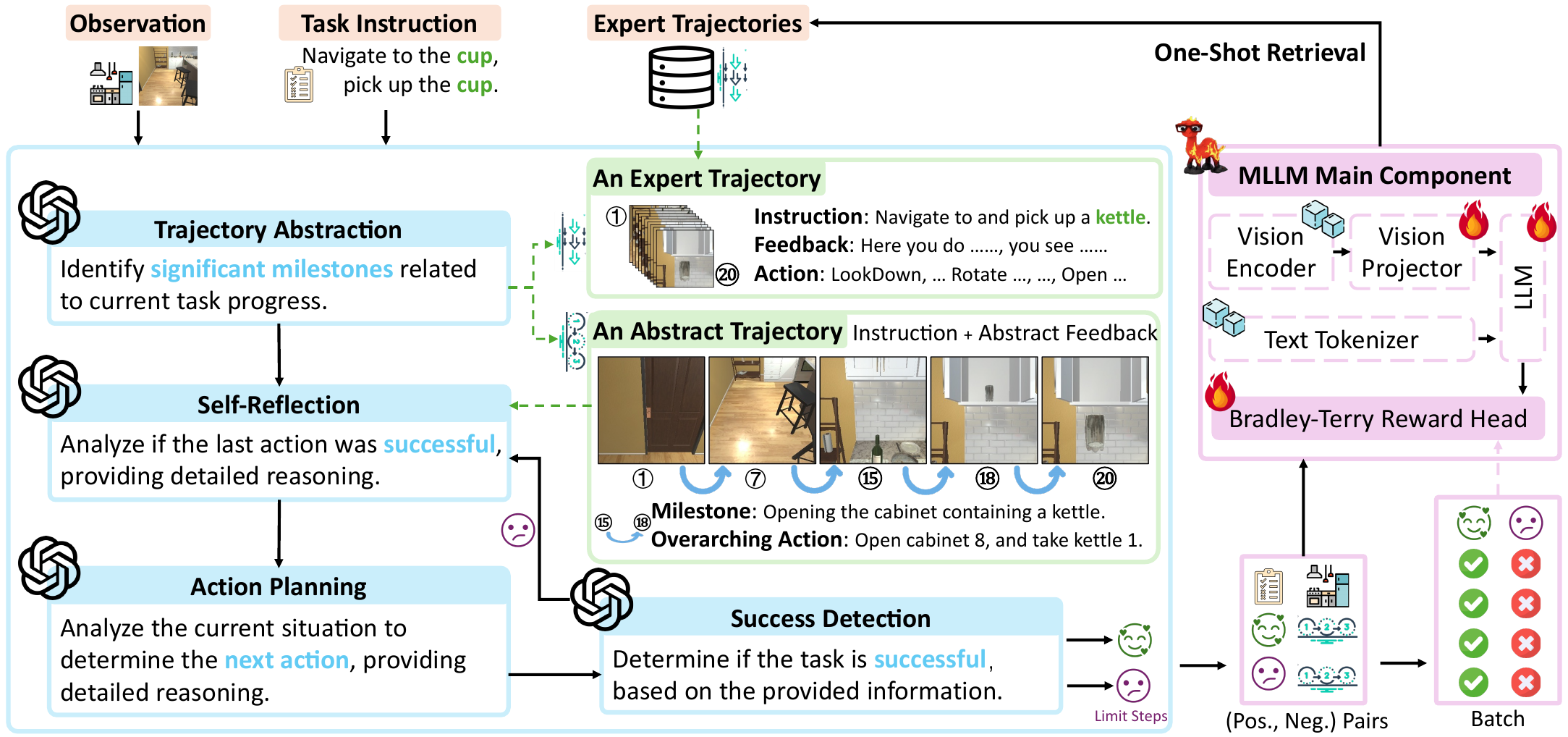}
    \caption{Overview of \projname. Our approach interactively learns a multimodal retriever to score expert trajectories and retrieve most effective trajectory to guide an agent in novel situations. By considering trajectories with higher success rates as positive samples and those with lower success rates as negative trajectories, we obtain the preference pairs, which are used to fine-tune an MLLM retriever to score trajectory effectiveness for a specific task. %Detailed information of \projname's data format can be found in Appendix \ref{sec:data_format}.
    }
    \label{fig:method}
    % \vspace{-2mm}
\end{figure}    

\subsection{Problem Formulation}

In this study, we investigate interactions of a retrieval-augmented MLLM agent with an environment to complete embodied tasks drawn from a specific distribution. 
Figure \ref{fig:method} provides an overview of our \projnameALL\ approach -- \projname. The agent is assigned a task instruction $\ell^c$ sampled from the task instruction distribution $p(\ell)$, and operates over a finite horizon $H$. 
At each timestep $t \in {1, 2, ..., H}$, the agent selects an action $a_t$ from the action space $A$ based on the current observation $o_t^c$ from the observation space $O$ and a reference trajectory $\tau^e$ retrieved from the expert trajectory memory $\mathcal{M}$.
The memory contains a set of multimodal expert trajectories $\mathcal{M} = \{\tau_1^e, \tau_2^e, ..., \tau_n^e\}$, where each trajectory $\tau_i^e$ includes task instructions $\ell_i^e$ sampled from the same task instruction distribution $p(\ell)$. The trajectory $\tau_i^e$ also contains observation sequences $\Vec{o _i^e} = \{o_{i_1}^e, o_{i_2}^e, ..., o_{i_H}^e\}$, and action sequences $\Vec{a_i^e} = \{a_{i_1}^e, a_{i_2}^e, ..., a_{i_H}^e\}$.

The agent follows a frozen policy $\pi(a|\ell^c, \tau^e, o^c)$, implemented as a Multimodal Large Language Model (MLLM), and the reference trajectory $\tau$ plays a significant role in grounding the agent within the embodied environment, supporting task accomplishment. 
This reference trajectory is retrieved through an MLLM retriever $q_\theta$.

% The training process fine-tunes an MLLM Retriever model $q_\theta$ within the training task distribution $p_\textrm{train}(\ell)$. 
We fine-tune our MLLM retriever on training task distribution $p_\textrm{train}(\ell)$ and evaluate its performance on test task distribution $p_\textrm{test}(\ell)$, which has no overlap with training tasks. %$p_\textrm{train}(\ell)$.
% We collect expert trajectories within the training task distribution to construct training memory $\mathcal{M^\textrm{train}}$, and sample trajectories from training memory as prompt for MLLM agent to complete training tasks. After that, we collect the induced success rate of these trajectories, and organize these trajectories into preference data, which is used to fine-tune an MLLM retriever.
This retriever aims to identify and retrieve a trajectory $\tau^e$ from the expert memory pool that is most effective for the current task $\ell^c$, \ie which can help ground the MLLM agent with a specific embodied task and enable its effective completion. 
It is worth noting that we have different memories $\mathcal{M^\textrm{train}}$ and $\mathcal{M^\textrm{test}}$, which corresponds to training tasks and test tasks, respectively.

\subsection{Memory}
\label{subsec:memory}

% To train our MLLM retriever, we 
To enable trajectory retrieval for task execution and fine-tuning our MLLM retriever, we first construct memory databases containing expert trajectories from previous successful executions for tasks both from $p_\textrm{train}(\ell)$ and $p_\textrm{test}(\ell)$. For each trajectory, 
% denoted as 
$\tau _i^e = \{\ell _i^e, \Vec{o _i^e}, \Vec{a _i^e}\}$,  
represents a task $\ell _i^e$ completed in $H_i$ steps and comprises the sequence of observations $\Vec{o _i^e}$, and corresponding actions $\Vec{a _i^e}$.

In multi-modal environments, such as \aithor~\citep{kolve2017ai2} and \legent~\citep{cheng2024legent}, each timestep observation $o_i$ includes an egocentric image. Moreover, in the \aithor\ environment, we assign numerical IDs (e.g. Cup 1) to all objects in the current visible field of view to identify target objects for interaction. These IDs appear in the environment feedback output in natural language, which is also part of the observation. %\borje{move the ID detail to settings?}

% The trajectories are collected by having the agent attempt the tasks and saving streams of successful episodes. The episodic trajectories capture the steps needed to complete the tasks. Storing these examples allows the agent to leverage past experience when facing new task instances.

Expert trajectories are collected via a planner-based method~\citep{hoffmann2001ff}. Storing these trajectories allows the agent to later leverage past experience when facing new task instances. Trajectory data is collected independently for the training and test sets. For each task, we initialized a task instance and used a planner-based method to collect expert trajectory data. It is worth mentioning that the initialization position and orientation in each task is randomly chosen.
% but the corresponding environment layout is fixed.

Since each task is directly corresponds to one unique task in the task set, the size of our memory used for experiments is relatively small.  
For example, in the experiments performed in the \legent\ environment, the memory pool consisted of 40 trajectories for training, and distinct 32 trajectories during testing. More details are available in the experimental settings (Section \ref{subsec:exp_setup}).

% To initialize memory, the number of collected trajectories is directly related to the number of unique tasks in the task set, which are easily obtained. For example, in the experiments performed in the \legent\ environment, the memory pool consisted of 40 trajectories for training, and distinct 32 trajectories during testing. More details are available in the experimental settings (Section \ref{subsec:legent}).

\subsection{Multimodal Retriever}
\label{subsec: multimodal retriever}

The core of \projname\ is the innovative use of interactive learning to train the trajectory retriever. For an embodied task, different trajectories stored in memory can be provided as references to the MLLM agent, leading to varying effects on the completion of the current task, depending on the degree of grounding with the environment they provide. Even if a trajectory has text instruction similar to the current task or an image sequence similar to the initial egocentric observation of the task, it does not guarantee that this trajectory can provide effective grounding. This is due to plain similarity alone not being able to reflect the effectiveness of the trajectory for the embodied task. 
For instance, a failed trajectory for a related task could have high textual and visual similarity to the target task.

In order to retrieve the trajectory that can provide the most benefit (\ie effective grounding) for the current task from the trajectory memory, we propose an interactive learning method for the MLLM retriever.
Specifically, for each task in training set, we sample $K$ trajectories from the training memory $\mathcal{M^\textrm{train}}$, and feed them as prompt for MLLM agent to execute the embodied task respectively. After that, based on the induced success rates of task execution, we can get the effectiveness of each trajectory for the embodied task. 
% We can then obtain a list of partial order based on the comparisons the success rate, which produces $\binom{K}{2}$ pairs through pairwise comparison, where the trajectory with higher success rate will be treated as positive item and the trajectory with lower success rate will be treated as negative item. 
We can then obtain a partial order list based on success rate comparisons, producing $\binom{K}{2}$ pairs through pairwise comparison, where the trajectory with a higher success rate is treated as the positive item, and the one with a lower success rate as the negative item.
In this way, these preference pairs from interactive feedback are arranged as a positive-negative pair dataset $D$, which we use to fine-tune the MLLM according to the Bradley-Terry \citep{bradley1952rank} Reward Modeling loss to enhance its critiquing ability, as in Equation \ref{eq:criticLoss}:
\begin{equation}
\mathcal{L}(\theta) = - \mathbb{E}_{(\ell _1^c, o_1^c, \mathcal A(\tau _w^e), \mathcal A(\tau _l^e)) \sim D} \left[ \log \left( \sigma \left( q_\theta(\ell ^c, o_1^c, \mathcal A(\tau _w^e)) - q_\theta(\ell ^c, o_1^c, \mathcal A(\tau _l^e)) \right) \right) \right].
\label{eq:criticLoss}
\end{equation}

% where positive and negative sample pairs are generated based on the performance differences (\eg success rates) of multiple trajectories for the same task. 

% During this process, the MLLM is provided with the task instructions, the initial observation, and a reference trajectory. Its output is a scalar value representing the effectiveness of the given trajectory in completing the task.

% In particular, we use is LLaVA-7B~\citep{liu2024visual} as base MLLM and add a Bradley-Terry reward head based on hidden states of base model output. The reward head is a one-layer MLP that takes as input the last token in the hidden state and outputs a scalar reward. The input to the MLLM retriever includes the initial egocentric observation $I _0$, end image $I _T$, and task instruction $\ell _i$ in trajectory $\tau _i$; current state $S _0$ and current task instruction $\ell _0$; and a prompt for it to judge the affordance of this trajectory for the current task and state. Upon inference, the MLLM outputs the score of the trajectory, which indicates its effectiveness for the current task. We select the highest score trajectory as reference for the embodied agent. 

In particular, we use LLaVA-7B~\citep{NEURIPS2023_6dcf277e}
as base MLLM and add a Bradley-Terry score head 
% \citep{bradley1952rank}
based on hidden states of base model output. The score head is a one-layer MLP that takes as input the last token in the hidden state and outputs a scalar score. The input to the MLLM retriever includes the trajectory abstraction result of expert trajectory $\tau _i^e$, $\mathcal A(\tau_i^e)$, current observation $o_1^c$, current task instruction $\ell^c$, and a prompt for it to judge the effectiveness of this trajectory for the current task and state. Upon inference, the MLLM outputs the score of the trajectory, which indicates its effectiveness for the current task. We select the trajectory with the highest score as the reference for the embodied agent to complete the current task.

\subsection{Trajectory Abstraction}
A complete multimodal trajectory often has dozens of timesteps, which correspond to dozens of observations, actions, and feedback, including redundant or irrelevant information. Furthermore, inputting all trajectory tokens, especially image tokens, to the MLLM retriever and agent will likely lead to confusing the models/agent or even exceeding their context windows. 

We thus use another MLLM (in our experiments, GPT-4o) to automatically create an abstract trajectory in zero-shot manner. The initial input to the MLLM is the trajectory $\tau^e$ (consisting of task instruction $\ell^e$, observation sequence $\Vec{o^e}$, and action sequence $\Vec{a^e}$) as well as the current task instruction $\ell ^c$, and we let the MLLM find whether each observation contained in the trajectory $\tau^e$ is helpful for the current task $\ell^c$. If it is considered to be useful, we will keep the observation into the resulting trajectory abstraction $\mathcal A(\tau ^e)$. 
% whether trajectory $\tau^e$ is helpful for the current task $\ell^c$ and keep it, if so. 

% Firstly we let the MLLM to try and fully comprehend the tasks accomplished in the given trajectory $\tau^e$, and then identify its significant milestones (\ie which steps in the trajectory are essential for accomplishing the task). These are points where important decisions are made, goals are achieved, or notable changes in the environment or state occur. \eg if the target object of current task instruction $\ell ^c$ appears in the trajectory feedback, then the point where it appears is also a significant milestone as it is strongly related to current task $\ell ^c$.
% \junpeng{Why here use e.g.?} \borje{this is just one example of important point}

To be more specific, we let the MLLM comprehend the tasks accomplished in the given trajectory $\tau^e$, and then identify important observations in the trajectory as milestones and preserve them into the resulting trajectory abstraction. 
% (\ie which steps in the trajectory are essential for accomplishing the task)
These milestones are steps that are essential for accomplishing the trajectory task of $\ell ^e$, such as steps where important decisions are made, goals are achieved, or notable changes in the environment or state occur. Also if the target object of current task instruction $\ell ^c$ appears in the trajectory feedback, then the step where it appears is also considered to be a significant milestone as it is strongly related to current task $\ell ^c$.
% These are points where important decisions are made, goals are achieved, or notable changes in the environment or state occur. Also if the target object of current task instruction $\ell ^c$ appears in the trajectory feedback, then the point where it appears is also a significant milestone as it is strongly related to current task $\ell ^c$. 
The milestone output format consists of: 1) a description of the milestone; 2) the corresponding image (\{image x\}); 3) the corresponding feedback (\{feedback x\}); and 4) the overarching actions taken between this milestone and the next one.

%\begin{compactitem}
%    \item The corresponding image (\textless image x\textgreater).
%    \item The corresponding feedback (\textless feedback x\textgreater).
%    \item The overarching actions taken between this milestone and the next one.
%\end{compactitem}

% The summarized trajectory manages to remove redundant information without affecting relevance for a given task, so that the agent can better receive the grounding information contained in the trajectory. In a quantitative manner, for the tasks in the test set, the input trajectories of trajectory abstraction have an average of 11.5097 steps, while the output milestones have an average count of 3.1316.

The summarized trajectory manages to remove redundant information without affecting relevance for the current task $\ell ^c$, so that the agent can better receive grounding information contained in the trajectory. Quantitatively, for the tasks in the test set, the input trajectories of trajectory abstraction have an average of 11.51 steps, while the output milestones have an average count of 3.13.

% We use zero-shot GPT-4o to automatically summarize the trajectory, divide the trajectory into several steps according to the key milestone and summarize the instructions of each one. We only need to store the key milestone and overarching actions, which greatly reduces the pressure of inputting too many tokens while maintaining the amount of information.

% For trajectory $\tau$, current task $\ell$, trajectory summarization will  

% \borje{Overarching action}

\section{Experiments}
\label{sec:experiments}

\subsection{Experimental Setup}
\label{subsec:exp_setup}

\begin{figure}[htbp]
    \centering
    % Left minipage for the images
    \begin{minipage}{0.356\textwidth}
        %\vspace{-10pt}
        \centering
        \subcaptionbox{\aithor}[\linewidth]{
            \includegraphics[width=0.45\linewidth]{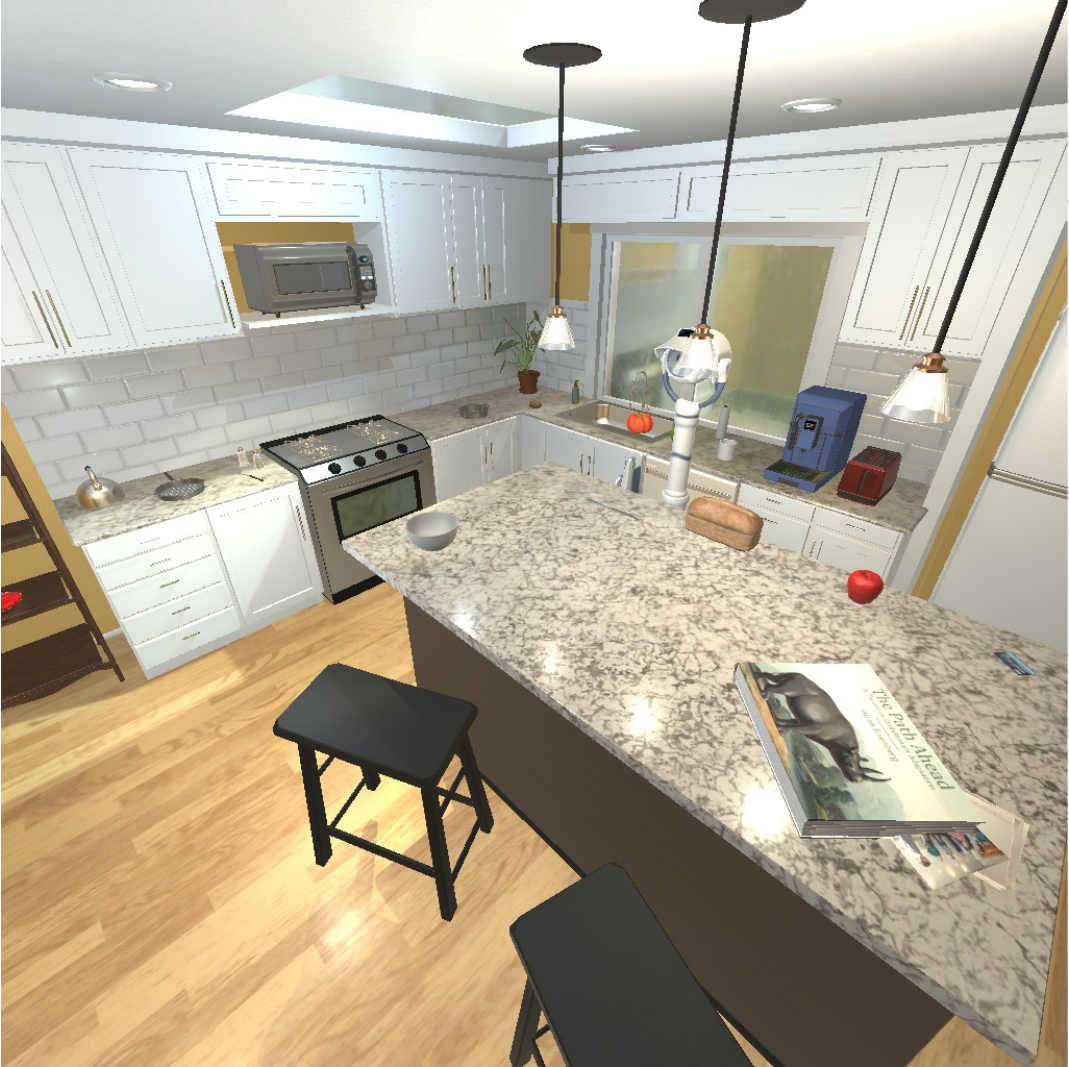}
            \includegraphics[width=0.45\linewidth]{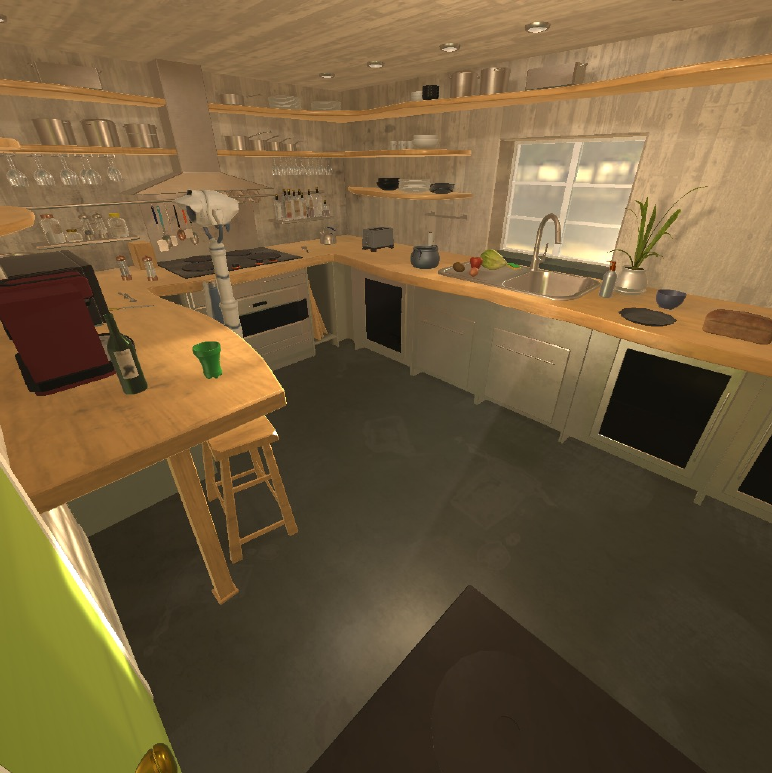}
        }
        
        \subcaptionbox{\legent}[\linewidth]{
            \includegraphics[width=0.45\linewidth]{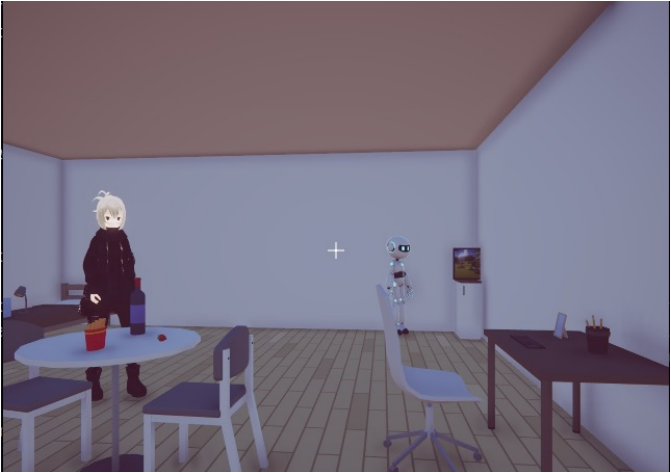}
            \includegraphics[width=0.45\linewidth]{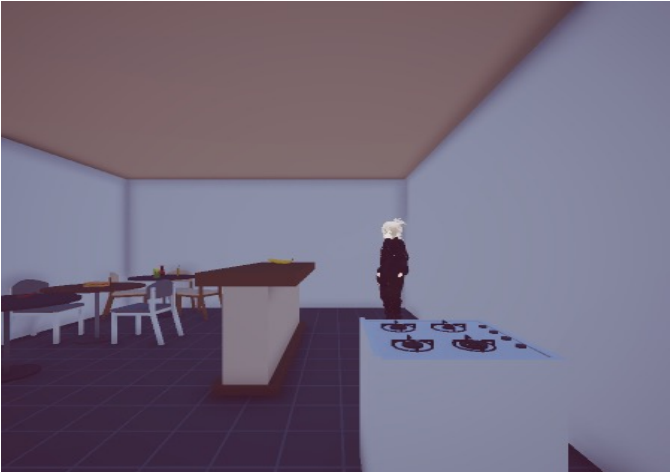}
        }
        \vspace{-5pt}
        \caption{Environment comparison.}
        \label{fig:environment_comparison}
    \end{minipage}%
    \hspace{3mm}
    % Right minipage for the table
    \begin{minipage}{0.57\textwidth}
        %\vspace{-2pt} % Adjust vertical spacing if needed
        \centering
        \captionof{table}{Environment complexity comparison. \aithor\ contains more than four times the number of interactive objects per scene, and more complex object hierarchy, compared to \legent. E.g., \aithor\ supports relationships such as ``inside" (e.g., inside a microwave (Figure \ref{subfig:Microwave}), or inside an open container, like a sink (Figure \ref{subfig:Sink})). }
        \vspace{-2mm} % Adjust spacing as needed
        \small
        \begin{tabular}{lrr}
            \toprule
             & \textbf{\aithor} & \textbf{\legent} \\ \midrule
            Avg. Objects/Scene & 47.30  & 11.13 \\ 
            Object Hierarchy & Inside, On & On \\ 
            Layout Complexity & High & Low \\ 
            Task decomposition & Yes & No \\ 
            Observation & Image, Feedback & Image \\\bottomrule
        \end{tabular}
        \label{tab:environment_comparison_table}
    \end{minipage}
\end{figure}

\subsubsection{Environments}
To validate the effectiveness of our method in various environments, we perform evaluations on multiple scenarios in two environments, \aithor~\citep{kolve2017ai2} and \legent~\citep{cheng2024legent}. %Both of which are multimodal environments, whose observations are egocentric images, and both make use of fine-grained control actions.
Unlike LLM agents, which use only text as input and simplify the action space by employing teleportation actions (\eg ``move directly to target"), we believe that MLLM agents should undertake more challenging tasks as they have access to visual input and are no longer limited to a `blind' mode of operation. Therefore, both target environments are multimodal, whose observations are egocentric images, and both make use of fine-grained control actions.

\textbf{\aithor} simulates embodied household tasks supporting natural language instructions and egocentric visual observations, where agents must navigate and interact with various household items within realistic 3D environments, including kitchens, living rooms, and other indoor spaces. it allows agents to perform fine-grained navigation actions including `move ahead', `turn left/right x degrees', `look up/down', and interactive atomic action including `pick up object A', `put object A on/in object B', `open/close object A', `toggle on/off object A'.

\textbf{LEGENT} is designed to imitate human activities and tasks in home environments, including cross-room navigation. The action space is similar to \aithor's, including fine-grained movement actions. It also includes a `speak' action, which sends a message to the user. 

Table \ref{tab:environment_comparison_table} and Figure \ref{fig:environment_comparison} show a comparison between the two environments. Besides having many more objects in its scenes, \aithor\ supports more complex object hierarchy relationships, such as ``inside" (\eg ``inside a cabinet", which require open/close interaction to complete a task, or ``a cup inside a sink", \ie an open container), not supported in  \legent~\footnote{Although there are cabinets in \legent, they do not need to be opened/closed to complete tasks.}.

%We integrated two benchmarks based on \aithor: ALFWorld and ALFRed, and built an environment setting that is more suitable for the MLLM agent.

\subsubsection{Task Settings}

Notably, in all tasks, the initial position and orientation of the agent is chosen randomly. Each task is tested 5 times to reduce the impact of random errors.

\textbf{\aithor}.
To better assess the fine-grained control ability of MLLM agents to complete real-world embodied tasks, we integrate characteristics of two \aithor-based benchmarks -- ALFWorld~\citep{ALFWorld20} and ALFRed~\citep{Shridhar_2020_CVPR} -- and built an environment setting that is more suitable for the MLLM agent. Since tasks are long-horizon, we follow the method in ALFRed and apply task decomposition to divide them into sub-tasks before execution. Each sub-task is then provided to the agent, and it determines sub-task success based on environmental feedback by itself (details in Appendix~\ref{appendix_task_decomposition} and ~\ref{appendix_success_detection}
). Once a sub-task is successfully completed, the agent proceeds to the next sub-task. Task types include \verb|pick_and_place|, \verb|pick_clean_then_place|, \verb|pick_cool_then_place|, and \verb|pick_heat_then_place|. Completing these tasks requires dozens of steps of navigation, as well as interaction with objects. There are 45 tasks comprising a total of 260 sub-tasks in training set, and 28 tasks including 158 sub-tasks in testing set. % We also construct a generalization test set to evaluate the performance in unseen tasks with unseen an task type, which contains 10 additional tasks.

\textbf{LEGENT}.
Tasks for the MLLM agent are categorized into two types: `Come Here' and `Where Is'. Each task is further divided into `One-room' and `Two-room' types, based on whether it requires traversing between rooms. In LEGENT, task decomposition is not performed as task instructions are simpler and do not contain combinations of sub-tasks; \ie the granularity of tasks is similar to that of sub-tasks in \aithor. To train the retriever, we use 40 tasks (10 tasks for each task type) and we use 32 tasks, also covering all task types, as test set. 

\subsubsection{Memory Construction}

Memory initialization follows the procedure described in \Cref{subsec:memory}; with randomized starting positions.

\textbf{\aithor}. Once trajectories are collected, we decompose them into sub-task trajectories (following the task decomposition procedure in ALFRed) and treat each sub-task level trajectory as a expert trajectory into memory. %, and agent retrieve trajectories when it face each sub-task. 
%Sub-tasks and trajectories are in the same level. 
Similar redundant trajectories are then filtered out, accounting for about one-third of total, resulting in a collection of 170 memory trajectories for the training memory and 118 trajectories for the testing memory.

\textbf{LEGENT}. As decomposition is not necessary due to the simpler tasks in this environment, the training memory is initialized with 40 trajectories, and the test memory with distinct 32 trajectories; one trajectory per task.
%In order to align the amount of training data for different environments, we collected more trajectories for the training set. As a result, we collect 400 trajectories for training set and 32 trajectories for testing set. Each time a task is started, the starting position will be initialized randomly. 

\subsubsection{Task Evaluation}

To evaluate the effectiveness of the retrieval-augmented embodied agents, we assess their performance using two metrics: Success Rate (\textbf{SR}) and Average Steps (\textbf{AS}).

Success Rate denotes the percentage of tasks attempts successfully completed by the agent. In \aithor\, it indicates the percentage of completed full tasks, and we additionally use \textbf{SR-Sub} to represent the percentage of completed sub-tasks.

Average Steps represents the average number of steps the agent takes to complete a task. In \aithor\, it indicates the number of steps to complete a full task, and \textbf{AS-Sub} represents the step average to complete a sub-task. Notably, for failed cases, their steps are counted as the step limit.

\subsubsection{Baselines}
It is worth noting that \projname\ is the first work to retrieve multimodal trajectories as references for embodied MLLM agents and let the agent directly output fine-grained control actions. We compare \projname\ against three baseline methods to explore the performance of our approach:

\textbf{Plain-Agent (PA)} is an embodied MLLM agent without making use of reference trajectories, \ie without any memory. In our experiments, we use GPT-4o (2024-05-13 version).

\textbf{LLaVA-Plain (LP)} is a pre-trained LLaVA with no modified head and no finetuning. We use the probability of special token generation to represent the score. Its input is the same as \projname, and it is prompted to output only Yes/No tokens, and the final score is calculated based on the probability of token generation (more details and limitations in Appendix~\ref{llava_plain}).  

\textbf{Similarity+LLaVA (SL)} is a reasonable retrieval+ranking approach. Such approach is common in the text retrieval field -- \eg \citet{sun2023chatgpt,sun2023instruction,dong2024rlhf} -- and it can take into account both similarity and effectiveness. We use similarity to choose the top-K candidate trajectories, and then choose the most likely effective one using a plain LLaVA model (\ie same as \textbf{LP}).

\textbf{RAP}~\citep{kagaya2024rap} performs retrieval based on plain similarity per modality and is the most similar setting in literature. It mainly targets text modality experiments in the ALFWorld~\citep{ALFWorld20} and WebShop~\citep{WebShop} (treated as text-only) environments, and simple tasks in the multimodal environments Franka-Kitchen~\citep{Franka-Kitchen} and Meta-World~\citep{Meta-World}.

\begin{table}[!t]
\centering
\caption{Performance comparison of different methods in \aithor.}
\vspace{-2mm}
\small
\begin{tabular}{lrrrrr}
\toprule
 & \PlainAgent & \textbf{LP} & \SimilarityLLaVA & \textbf{RAP} & \textbf{MART} \\
\midrule
\textbf{SR} $\uparrow$ & 0.18   & 0.26   & 0.24   & 0.22 & \textbf{0.40} \\
\textbf{SR-Sub} $\uparrow$ & 0.63  & 0.69   & 0.68 & 0.67  & \textbf{0.75} \\
\textbf{AS} $\downarrow$ & 159.66   & 144.18   & 147.48   & 147.03 & \textbf{123.19} \\
\textbf{AS-Sub} $\downarrow$ & 44.65   & 39.88   & 40.79    & 40.67   & \textbf{34.07} \\ %\textbf{AS-Sub} $\downarrow$ & 44.65   & 37.27    & 40.67   & \textbf{34.07} \\
% \midrule
% & \multicolumn{4}{c}{Generalization Task Set} \\ 
% \midrule
% \textbf{SR} $\uparrow$ & 0.05  & 0.33    & 0.30  & \textbf{0.60} \\
% \textbf{AS} $\downarrow$ &    &     &    &  \\
\bottomrule
\label{tab:ai2thor_results}
\end{tabular}
\end{table}

\begin{table}[!t]
\vspace{-4mm}
\centering
\caption{Performance comparison of different methods in \legent.}
\label{tab:legent_results}
\vspace{-2mm}
\small
\begin{tabular}{lrrrrr}
\toprule
& \PlainAgent & \textbf{LP} & \SimilarityLLaVA & \textbf{RAP} & \textbf{MART} \\
\midrule
\textbf{SR} $\uparrow$ & 0.70 & 0.69 & 0.75 & 0.75 & \textbf{0.87} \\
\textbf{AS} $\downarrow$ & 23.62 & 25.01 & 20.92 & 20.62 & \textbf{13.81}  \\
\bottomrule
\end{tabular}
\end{table}

\begin{table}[!t]
    \centering
    \caption{Ablation studies of \projname\ in the \aithor\ and \legent\ environments.}
    \label{tab:combined_ablation_results}
    \vspace{-2mm}
    \small
    \begin{tabular}{llrrr}
    \toprule
    \textbf{Environment} & \textbf{Metric} & \textbf{w/o Abstraction} & \textbf{Sim.+FTM} & \textbf{MART} \\
    \midrule
    \multirow{4}{*}{\textbf{\aithor}} &\textbf{SR} $\uparrow$ & 0.31 & 0.34 & \textbf{0.40} \\
     & \textbf{SR-Sub} $\uparrow$ & 0.73 & 0.74 & \textbf{0.75} \\
     & \textbf{AS} $\downarrow$ & 130.26 & 125.20 & \textbf{123.19} \\
     & \textbf{AS-Sub} $\downarrow$ & 36.03 & 34.63 & \textbf{34.07} \\
     \midrule
     \multirow{2}{*}{\textbf{\legent}} & \textbf{SR} $\uparrow$ & 0.77 & 0.77 & \textbf{0.87} \\
     & \textbf{AS} $\downarrow$ & 18.48 & 18.83 & \textbf{13.81} \\
    \bottomrule
    \end{tabular}
\end{table}

\subsection{\aithor}
\label{subsec:ai2thor}

We firstly demonstrate the effectiveness of \projname\ over test tasks in the \aithor\ environment. %, which is unseen for \projname. 
The experimental results demonstrate the effectiveness of our approach compared to the baselines. As shown in Table \ref{tab:ai2thor_results}, \projname\ surpasses all baselines over 10\% in Success Rate, and reaches best performance across all metrics. 

\subsection{\legent}
\label{subsec:legent}

% \legent\ is a multimodal embodied environment incompatible with LMM agents. The action space is similar with \aithor, including fine-grained movement and interaction actions. Besides, it includes `Speak' action, which is sending a message to the user. The observation of \legent\ is also egocentric images.

% Following the setup of \legent, the tasks for the MLLM agent are categorized into two types: `Come Here' and `Where Is'. Each task is further divided into One-room and Two-room based on whether it requires traversing between rooms.

% To train the retriever, we used 40 tasks, with 10 tasks for each task type. To evaluate the effectiveness of the approach, we used 32 tasks, covering all task types. Notably, in the test set, the scene of each task is randomly initialized. Each task was tested 5 times to reduce the impact of random errors. The expert trajectories are generated by planner-based method.

% \borje{generated scenes, to be released. maybe describe more in appendix}

We then conduct experiments in the \legent\ environment. This environment includes tasks involving crossing between rooms, thereby enriching the experimental space and demonstrating the effectiveness of our method. The experimental results, shown in Table \ref{tab:legent_results}, demonstrate \projname\ greatly surpasses all baselines in all four task types.
% especially in the Come-2room type, where \projname\ successfully avoids the drop in performance shown by \textbf{RAP}.

% \begin{table}[h!]
% \centering
% \begin{tabular}{lcccc}
% \toprule
%  & \textbf{w/o memory} & \textbf{LLaVA-plain} & \textbf{RAP} & \textbf{MART} \\
% \midrule
% Where-2room & 0.65   & 0.62   & 0.725  & \textbf{0.875} \\
% Where-1room & 0.78   & 0.82   & 0.7333 & \textbf{0.9111} \\
% Come-2room  & 0.625  & 0.6    & 0.4625  & \textbf{0.725} \\
% Come-1room  & 0.75   & 0.8    & 0.85   & \textbf{0.9778} \\
% \midrule
% \textbf{Average SR.} & 0.70125 & 0.71   & 0.6927 & \textbf{0.872225} \\
% \bottomrule
% \end{tabular}
% \end{table}

% \begin{table}[h!]
% \centering
% \caption{Performance comparison of different scenes in \legent.}
% \begin{tabular}{lcccc}
% \toprule
%  & \textbf{w/o memory} & \textbf{Simi+LLaVA-plain} & \textbf{RAP} & \textbf{MART} \\
% \midrule
% Where-2room & 0.65   & 0.6   & 0.725  & \textbf{0.875} \\
% Where-1room & 0.78   & 0.8   & 0.7333 & \textbf{0.9111} \\
% Come-2room  & 0.625  & 0.625    & 0.4625  & \textbf{0.725} \\
% Come-1room  & 0.75   & 0.9556    & 0.85   & \textbf{0.9778} \\
% \midrule
% \textbf{Average SR.} & 0.70125 & 0.74515   & 0.6927 & \textbf{0.872225} \\
% \bottomrule
% \label{tab:legent_results}
% \end{tabular}
% \end{table}

\subsection{Ablations}

In this section, we use a set of ablation studies to examine the contribution of key components in \projname. More specifically, we aim to answer the following questions.

\textbf{Q1.} Does Trajectory Abstraction indeed improve embodied agent performance?

We compare the full \projname\ approach and the ablation of removing its Trajectory Abstraction module (\textbf{w/o Abstraction}). As shown in Table \ref{tab:combined_ablation_results}, \projname\ consistently reaches the best results across settings. Even if in \aithor\ sub-tasks the average success rate is comparable, the improvement margin leads to a 9 percentage points improvement in full task success rate.

\textbf{Q2.} How does the \projname\ approach compare against a typical retrieve and rank approach, even if it uses a fine-tuned ranking model for trajectory usefulness?

Table~\ref{tab:combined_ablation_results} shows decomposing the unified \projname\ approach into separate retrieval and ranking steps (\ie similarity-based retrieval and \projname's MLLM as ranker -- \textbf{Sim.+FTM}) decreases success rates across settings.  It is also interesting to note that \textbf{Sim.+FTM} outperforms both \textbf{SL} and \textbf{RAP} (Tables~\ref{tab:ai2thor_results} and~\ref{tab:legent_results}), further illustrating \projname's trajectory utility scoring.

\begin{figure}[t]
    \centering
    % \vspace{-4mm}
    \includegraphics[width=.95\linewidth]{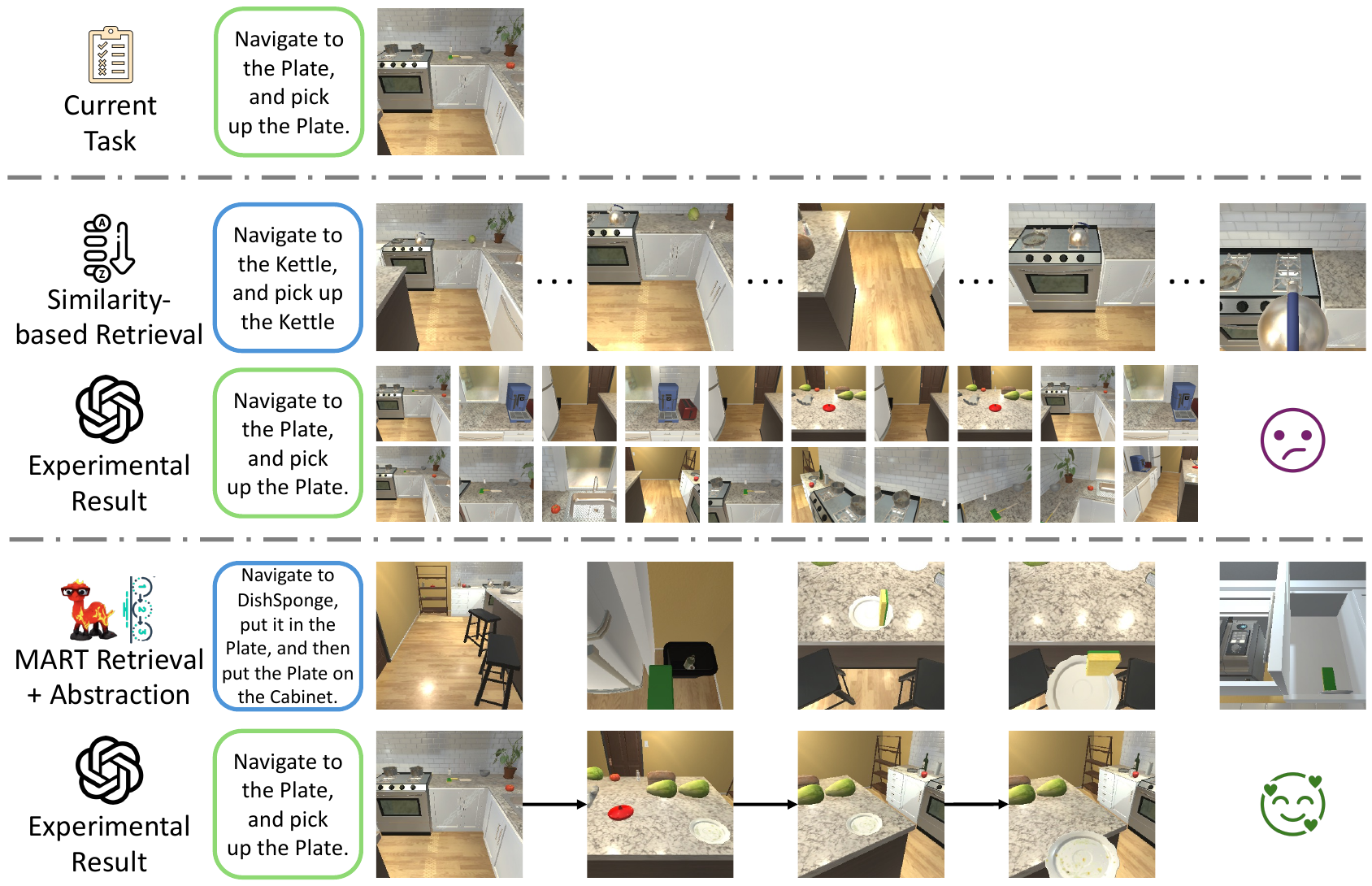}
    \caption{Comparison between similarity-based retriever and \projname.}
    \label{fig:case_study_1}
    % \vspace{-2mm}
\end{figure}

\subsection{Case Study}

% \begin{wrapfigure}{r}{0.6\textwidth}
%     \vspace{-10pt}
%     \centering
%     \includegraphics[width=\linewidth]{ICLR 2025 Template/images/Case_study_1st.pdf}
%     \caption{Case Study 1. Comparison between \projname and similarity-based retrieval method. Although the results retrieved by the similarity-based method have high similarity, they do not contain content that is useful for the current task. \projname effectively retrieves the trajectory containing the target object required by the current task and helps the agent complete the task efficiently. pure similarity-based retrieval method has many repeat states.}
%     \label{fig:case_study_1}
%     \vspace{-10pt}
% \end{wrapfigure}

% While the similarity-based method finds highly similar results, they often lack useful content. In contrast, \projname efficiently retrieves the required trajectory, helping the agent complete the task without repeated states.

We present two case studies for more in-depth discussion of \projname's capabilities handling challenges of the MLLM agent setting. %of highlighting its improved trajectory retrieval and the effectiveness of Trajectory Abstraction.
The first case (Figure \ref{fig:case_study_1}) demonstrates the effective handling of a very long-horizon trajectory. Given a 73-step task trajectory -- ``navigate to DishSponge, put it in the Plate, and place the Plate on the Cabinet" -- Trajectory Abstraction identifies 5 key milestones. \projname\ achieves an 80\% success rate with an average of 28 steps, while both the similarity-based method and the agent without memory reach only 40\% success rate, averaging 69.6 and 74.6 steps, respectively. We provide more detailed and balanced case studies in Appendix~\ref{sec:detailed case study}.

\begin{figure}[ht]
    \centering
    % \vspace{-4mm}
    \includegraphics[width=.95\linewidth]{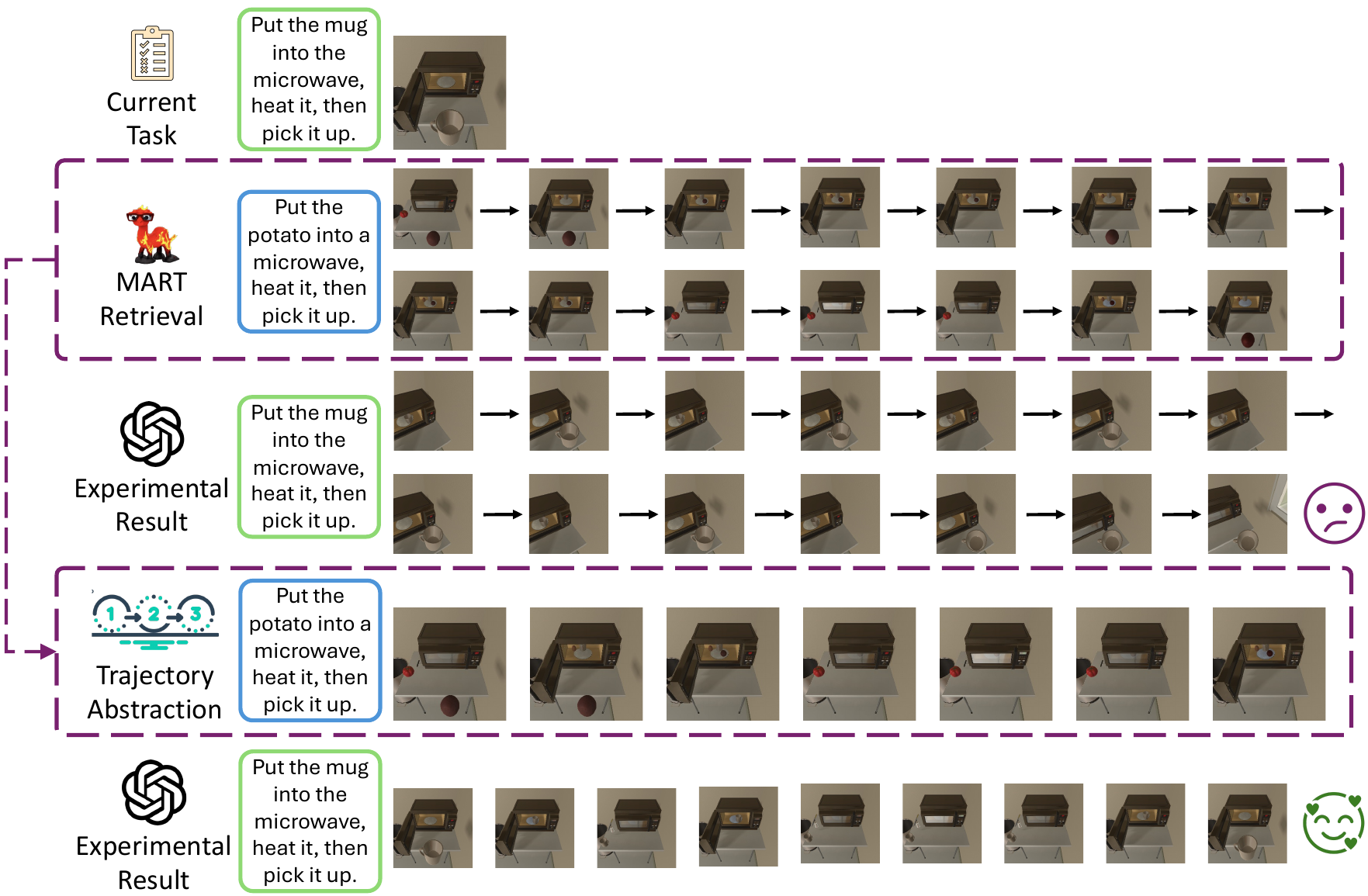}
    \caption{Showcase of the significance of the Trajectory Abstraction mechanism.}
    \label{fig:case_study_2}
    % \vspace{-2mm}
\end{figure}

The second case (Figure \ref{fig:case_study_2}) shows how \projname\ extracts implicit rules for long sequence tasks. For the task ``put the Potato into the microwave, heat it, and pick it up”, Trajectory Abstraction analyzes the agent's actions (including exploration, attempts, and success) and generates an abstract set of inferred rules, pruning non-contributory and redundant actions, reducing 13 transitions to just 6. \projname\ achieves an 80\% success rate with an average of 30.2 steps, while other methods have a 0\% success rate.

% \begin{wrapfigure}{r}{0.6\textwidth}
%     \vspace{-10pt}
%     \centering
%     \includegraphics[width=\linewidth]{ICLR 2025 Template/images/Case_study_2nd.pdf}
%     \caption{Case Study 2. Showcase the significance of Trajectory Abstraction.}
%     \label{fig:case_study_2}
%     \vspace{-10pt}
% \end{wrapfigure}

% \section{Conclusion and Limitations}
\section{Conclusion}
\label{sec:conclusion}

We propose \projname, a new paradigm for trajectory retrieval incorporating interactive learning, to enhance embodied agents' performance by providing them with task-relevant trajectory data. Our approach utilizes interaction-based feedback to identify the most effective trajectories, and constructs preference pairs based on the comparisons between trajectories.
% task success rate.
An MLLM retriever is fine-tuned through these preference pairs, effectively prioritizing the trajectories that improve task performance. 
% where an MLLM retriever is fine-tuned to score those based on task success rates, effectively prioritizing the ones that improve task performance. 
We also introduce Trajectory Abstraction in \projname, a novel mechanism that leverages MLLMs’ summarization capabilities, to abstract trajectories, \ie reduce the required number of tokens to represent them, while preserving key information and enabling agents to better understand relevant information. Experimental results in different environments demonstrate that our method significantly enhances task success rates in unseen tasks, compared to multiple baselines. This work helps bridge the gap between general-purpose MLLMs and the specific requirements of embodied tasks, offering a new paradigm for multimodal trajectory retrieval for embodied agents.

\section{Discussions and Future Works}
\label{sec:future_work}

Our study has a few limitations. First, the MLLM's restricted context window limits its ability to process multiple images simultaneously, restricting us to one-shot learning with single trajectory inputs. Future work will explore few-shot learning to combine skills from multiple trajectories for enhanced performance in complex tasks.

Second, the lack of a detailed ablation study will be addressed by evaluating each component's contribution, including self-reflection mechanisms, prompt designs tailored to specific functions in action planning, and retrievers based on diverse base models.
% function-specific prompt designs, and diverse retriever models, alongside comparisons with other MLLM agents.

Third, to ensure fair comparisons, we will construct large-scale, high-quality datasets specific to the household domain and fine-tune general feature extractors on these datasets to provide a more fair and direct comparison with similarity retrieval methods.

% fine-tune similarity-based retrieval methods on large-scale, domain-specific datasets, as current baselines are not optimized for our domain.

% Fourth, model transfer stability across environments within the same domain needs validation. We will train retrievers for household, web, and sandbox game domains to assess generalization in novel scenarios for each domain, considering differences in task nature, operational frequency, object morphology, and textual complexity.

% Lastly, our experiments are currently limited to the household domain. Future work will extend to web and sandbox game domains by collecting interaction data and training domain-specific retrievers, then evaluating generalization in unseen environments within each domain.

Fourth, model transfer stability across environments within the same domain needs validation. There are significant differences when comparing the household domain with the open-ended sandbox game domain and the web domain, including variations in task nature, operational frequency, object morphology, and textual input complexity. Therefore, we will train retrievers for the web, open-ended sandbox game, and household domains, respectively, to evaluate their generalization capability in novel environments and scenarios within each domain.

Lastly, the experimental scenarios in this study are somewhat limited, as we have only conducted experiments in the household domain. In future work, we plan to extend our approach to the web and open-ended sandbox game domains by collecting preference data through interactions and training retrievers for each domain. Subsequently, we will evaluate the generalization capabilities within the same domain, but in unseen environments and scenarios.

\subsubsection*{Acknowledgments}

This work was supported by NSFC under Grant 62450001 and 62476008. The authors would like to thank the anonymous reviewers for their valuable comments and advice.

%\clearpage
\bibliography{iclr2025_conference}
\bibliographystyle{iclr2025_conference}
\clearpage

\appendix
\section{Implementation details}
In this section, we provide more implementation details about the model, training process and implementation pipeline.

\subsection{Model and training details}
\label{subsec: model and training details}
We transform a generative language model (MLLM) into a trajectory scoring model by replacing the language model head with a Bradley-Terry score head. In particular, both the original language model head and our proposed Bradley-Terry scoring head are single-layer MLPs. However, there are notable differences: the language model head processes all hidden states as input and generates a probability distribution over the vocabulary for each token, facilitating token sequence generation through sampling. In contrast, our Bradley-Terry scoring head relies solely on the last non-zero hidden state as input and outputs a single floating-point score. Using this approach, our model generates only one new token at a time (i.e., by setting `max\_new\_tokens' to 1). In comparison to conventional MLLM training, which generates hundreds or even thousands of new tokens per iteration, our model's training is significantly more computationally efficient. All the code for the MLLM retriever model, training process, benchmark tasks and simulator modifications is available at \href{https://github.com/PKU-RL/MART}{https://github.com/PKU-RL/MART}.
% \textbf{All code, including the model, training process, benchmark tasks, and simulator, will be released} upon acceptance.

During training, we firstly fine-tune LLaVA to enable it to understand multiple images, details in \ref{appendix: multi-LLaVA}. After that, we replace the language model head with the Bradley-Terry score head, and fine-tune the model with lora. The parameter settings are listed in \ref{tab:hyperparameters}.

\subsection{Details of implementation pipeline}

In this section, we provide more details about the whole pipeline of implementation, including the data collection and retrieval in downstream tasks. The implementation pipeline is as follows:

\begin{enumerate}
    \item Construct memory databases containing expert trajectories via the planner-based method, details in \ref{subsec:memory}. 
    % We will release the trajectory collection code and guidelines along with the simulator code.
    \item Collect the pairwise comparison data via interactive feedback to train the retriever model. 
    \begin{enumerate}
        \item Specifically, for each task in the training set, we sample $K$ trajectories from the training memory $\mathcal{M^\textrm{train}}$, and feed them as prompts for MLLM agent to execute the embodied task respectively. 
        \item After that, based on the induced success rates of task execution, we can get the effectiveness of each trajectory for the embodied task. We can then obtain a partial order list based on success rate comparisons, producing $\binom{K}{2}$ pairs through pairwise comparison, where the trajectory with a higher success rate is treated as the positive item, and the one with a lower success rate as the negative item. 
        \item In this way, these preference pairs from interactive feedback are arranged as a positive-negative pair dataset $D$, which we use to fine-tune the MLLM according to the Bradley-Terry \citep{bradley1952rank} Reward Modeling loss to enhance its critiquing ability, as in Equation \ref{eq:criticLoss}. Details are listed in Appendix~\ref{subsec: multimodal retriever} line 249-259.
    \end{enumerate}
    \item Train the modified retriever model using the preference data. Details are listed in \ref{subsec: model and training details}.
    \item Evaluation on unseen tasks. For each unseen task, we first retrieve trajectory with highest score for current task. Then the retrieved trajectory will be simplified through Trajectory Abstraction module. After that, the MLLM agent will execute the task with the help of abstraction of retrieved trajectory. Details are listed in Algorithm \ref{algorithm:main}.
\end{enumerate}

\section{Extension of related works}

\subsection{Embodied Grounding}

Grounding is a critical challenge in embodied agents, referring to the alignment between the agent and its environment. 

The grounding problem can be categorized into visual grounding and embodied grounding. Visual grounding~\citep{Lai_2024_CVPR, kazemzadeh2014referitgame, nagaraja2016modeling} addresses the problem at the perception level by identifying the most relevant object or region in an image based on a language query, whereas embodied grounding focuses on the effects of actions on environmental dynamics and how an agent generates action sequences to accomplish a given task.
Approaches to addressing embodied grounding can be categorized into the following types:

\textbf{1. RL: }
Reinforcement learning (RL) trains an agent’s policy through interaction with the environment, making the agent inherently grounded in the environment, such as PPO~\citep{schulman2017proximal} and SAC~\citep{haarnoja2018soft}. However, RL typically requires extensive interaction with environments and often suffers from instability, making it unsuitable for MLLMs.

\textbf{2. VLA:}
These methods focus on fine-tuning vision-language models (VLMs) using expert datasets collected from embodied environments, such as PaLM-E~\citep{driess2023palm} and RT-2~\citep{brohan2023rt}. These methods demand a significant amount of high-quality trajectory data for training.

\textbf{3. LLM as Planner:}
These methods leverage Large Language Models (LLMs) or Multimodal Large Language Models (MLLMs) to generate high-level plans, which are then translated into executable action sequences by low-level controllers, such as LLM-Planner~\citep{song2023llmplanner} and P-RAG~\citep{xu2024p}. A key limitation of these methods is their reliance on a predefined skill library, which restricts the scope of the agent’s capabilities. Besides, acquiring a skill library might require additional RL or IL training or prior knowledge about the environment~\citep{lifshitz2023steve, yuan2024pre}.

\textbf{4. Retrieval-Augmented MLLM Agent:} 
This category involves integrating task trajectory data into the prompts provided to MLLMs, such as RAP~\citep{kagaya2024rap}. These trajectory data, rich in grounding information about the environment, enable agents to perform tasks effectively. Retrieval-augmented methods usually demonstrate greater sample efficiency compared to RL and VLA, thanks to the use of an explicit memory buffer. Our work falls into this category.

\subsection{Multi-modal Information Retrieval}

Recent advances in multimodal retrieval have developed various methods for encoding, fusing, and measuring similarities across different modalities. ViLT \citep{kim2021vilt} directly embeds image patches with text using a Transformer, while ALIGN \citep{li2021align} and MURAL \citep{jain2021mural} use dual-encoder architectures with EfficientNet and BERT to align modalities through contrastive learning. IMAGEBIND \citep{girdhar2023imagebind} extends this by creating joint embeddings for six modalities, using ViT and Transformer models. Furthermore, \citep{changpinyo2021telling} integrates users' mouse trace interactions for refined image retrieval, while ReViz \citep{luo2023end} employs advanced encoding mechanisms for visual question answering.
Building on these encoding strategies, retrieval augmentation further enhances multimodal generation (\eg RAG~\citep{lewis2020retrieval}). 
% FLMR , RA-CM3, UniRAG, and MuRAG all leverage retrieval to improve tasks like visual question answering and image captioning through late interaction and fusion techniques. 
FLMR \citep{lin2023finegrained} addresses RA-VQA limitations by combining multi-dimensional embeddings from ColBERTv2 and ViT-based models for accurate knowledge retrieval. 
% FLMR \citep{lin2023finegrained}  addresses RA-VQA limitations by utilizing a late-interaction mechanism that combines multi-dimensional embeddings derived from ColBERTv2, Oscar, VinVL, and ViT for accurate knowledge retrieval.
RA-CM3 \citep{yasunaga2023retrieval} enhances image captioning and text-to-image generation by using a pre-trained CLIP model to augment inputs for a CM3 Transformer. Similarly, UniRAG \citep{sharifymoghaddam2024unirag} integrates retrievals using UniIR’s \citep{wei2023uniir} CLIP Score Fusion and BLIP Feature Fusion, improving performance in MLLMs like LLaVA. Lastly, MuRAG \citep{chen2022murag} introduces a retrieval-augmented transformer for KB-VQA, employing T5 and ViT for multimodal encoding and retrieval from a large-scale memory bank.
In contrast, our method prioritizes the effectiveness of retrieved information by employing interactive learning, ensuring that the information contributes directly to task completion.

\section{Extension Experiments}

\begin{figure}[ht]
  \centering
  \begin{subfigure}{0.19\textwidth}
    \centering
    \includegraphics[width=\linewidth]{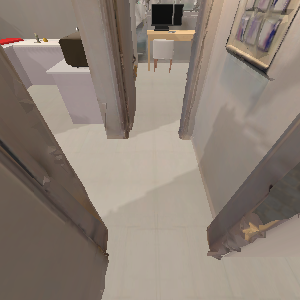}
    % \caption{.} 
    \label{subfig:realfred_3}
  \end{subfigure}
  \begin{subfigure}{0.19\textwidth}
    \centering
    \includegraphics[width=\linewidth]{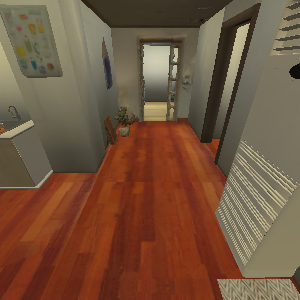}
    % \caption{.} 
    \label{subfig:realfred_1}
  \end{subfigure}
  \begin{subfigure}{0.19\textwidth}
    \centering
    \includegraphics[width=\linewidth]{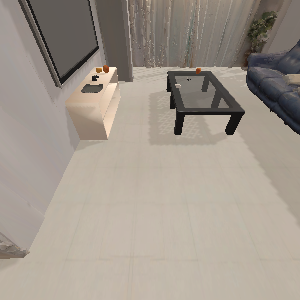}
    % \caption{.} 
    \label{subfig:realfred_4}
  \end{subfigure}
  \begin{subfigure}{0.19\textwidth}
    \centering
    \includegraphics[width=\linewidth]{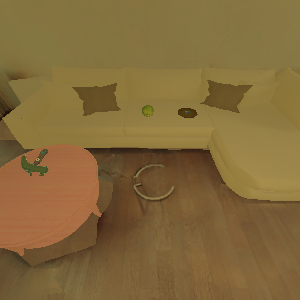}
    % \caption{.} 
    \label{subfig:realfred_2}
  \end{subfigure}
  \begin{subfigure}{0.19\textwidth}
    \centering
    \includegraphics[width=\linewidth]{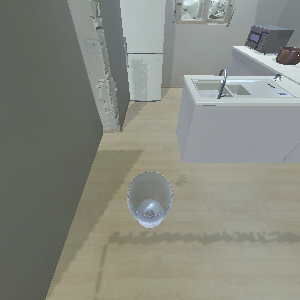}
    % \caption{.} 
    \label{subfig:realfred_5}
  \end{subfigure}

  \caption{Image examples of ReALFRED.} 
  \label{fig:realfred}
\end{figure}

We also evaluate our method on ReALFRED~\citep{kim2025realfred} environments, which provides realistic 3D-captured and multi-room scenes, as shown in Figure \ref{fig:realfred}. The action space is similar with AI2-THOR, including fine-grained movement actions, e.g. `move ahead', `turn left/right x degrees', `look up/down', and interactive atomic action including `pick up object A', `put object A on/in object B', `open/close object A', `toggle on/off object A'. Since completing tasks often involves navigating across rooms, and the scenes within these rooms closely resemble real-world environments, this setting offers a \textbf{more diverse and challenging scenario}. The chosen task type is `pick\_and\_place', which requires the agent to first navigate to the target object, pick it up, and then transport it to the designated location for placement. There are 30 tasks comprising a total 60 sub-tasks in training set, and 20 tasks including 40 sub-tasks in testing set.

The experimental results demonstrate the effectiveness of our approach compared to the baselines. As shown in Table \ref{tab:realfred_results}, \projname\ surpasses all baselines 10\% in Success Rate, and reaches best performance across all metrics. 

\begin{table}[ht]
\centering
\caption{Performance comparison of different methods in ReALFRED.}
\vspace{-2mm}
\begin{tabular}{lrrrrr}
\toprule
 & \PlainAgent & \textbf{LP} & \SimilarityLLaVA & \textbf{RAP} & \textbf{MART} \\
\midrule
\textbf{SR} $\uparrow$ & 0.25   & 0.20   & 0.26   & 0.27 & \textbf{0.37} \\
\textbf{SR-Sub} $\uparrow$ & 0.50  & 0.44   & 0.52 & 0.53  & \textbf{0.58} \\
\textbf{AS} $\downarrow$ & 101.70   & 87.93   & 89.79   & 87.89 & \textbf{78.48} \\
\textbf{AS-Sub} $\downarrow$ & 28.43   & 24.32   & 24.84    & 24.31   & \textbf{21.71} \\ 
\bottomrule
\label{tab:realfred_results}
\end{tabular}
\end{table}

\section{Detailed Case Study}
\label{sec:detailed case study}
We present more detailed case studies, encompassing both success and failure cases for each method, along with simplified reasoning processes for clarity.

\begin{figure}[ht]
    \centering
    % \vspace{-4mm}
    \includegraphics[width=\linewidth]{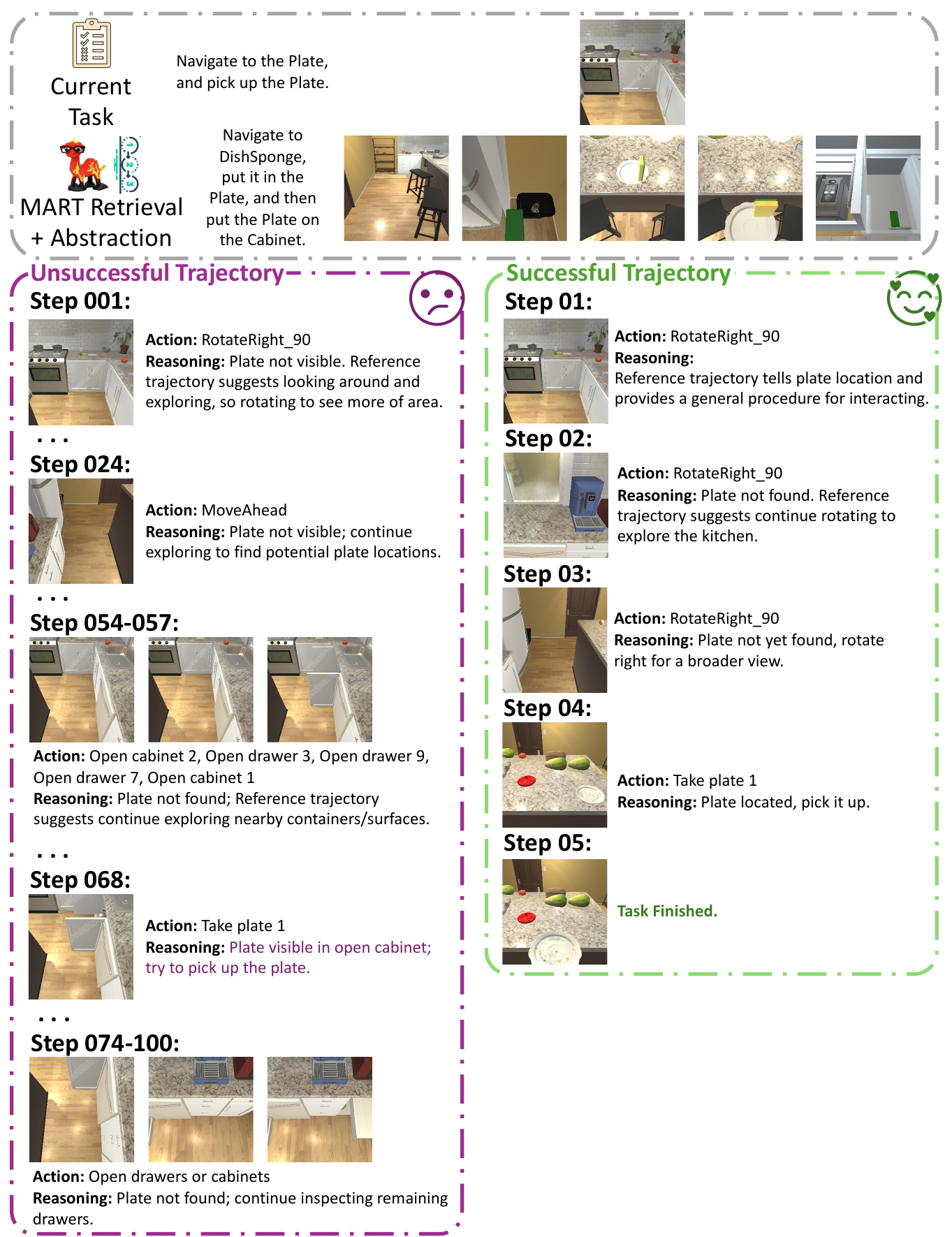}
    \caption{Comparison between unsuccessful and successful trajectory with \projname.}
    \label{fig:case_study_MART}
    % \vspace{-2mm}
\end{figure}  

\begin{figure}[ht]
    \centering
    % \vspace{-4mm}
    \includegraphics[width=\linewidth]{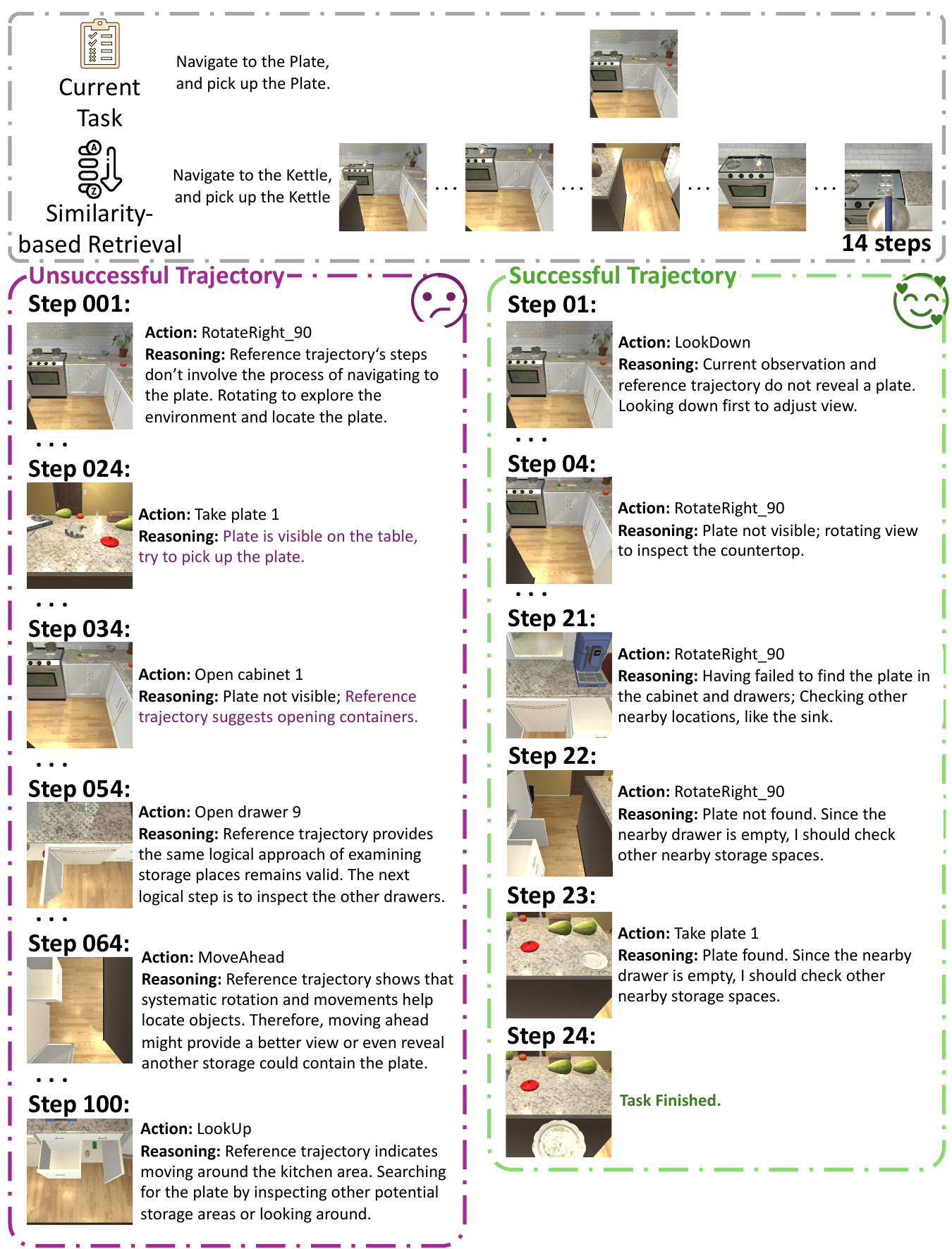}
    \caption{Comparison between unsuccessful and successful trajectory with similarity-based retrieval. }
    \label{fig:case_study_Sim}
    % \vspace{-2mm}
\end{figure}  

% In this first detailed case study, MART Retriever 检索出了包含target object位置的轨迹，而Trajectory Abstraction effectively compressed a trajectory of 73 steps into only 5 significant milestones while preserving crucial information.
% For the successful trial, the agent knows the location of the target object, the plate is on the table, through the retrieved trajectory, and after exploring, it finds the target object and succeeds the task.
% For the unsuccessful trial, the agent 在explore过程中出现了mistakes（紫色文字），直到达到步数上限也没有完成任务。

% In the second detailed case study, similarity-based retriever 检索出了很相似但没有提供有效信息的轨迹，导致agent只能自行探索。偶然地agent成功地找到了target object的位置，但总体成功率较低。

% This abstraction significantly improved the agent's task performance. Under identical conditions across other modules, the MART framework achieved a success rate of 80\% after five validations, compared to a 40\% success rate achieved by the similarity-based method.

In the first detailed case, as shown in Figure~\ref{fig:case_study_MART}, the MART Retriever successfully retrieved a trajectory containing the target object's location, while Trajectory Abstraction effectively compressed a 73-step trajectory into only 5 significant milestones, preserving crucial information.
For the successful trial, the agent identified the target object's location (the plate on the table) through the retrieved trajectory. After some exploration, it successfully found the target object and completed the task.
For the unsuccessful trial, during exploration, the agent made mistakes (highlighted in purple) and failed to complete the task within the step limit.

In the second detailed case, as shown in Figure~\ref{fig:case_study_Sim}, the similarity-based retriever retrieved a trajectory that appeared similar but lacked useful information. As a result, the agent had to explore independently. By chance, the agent located the target object, but the overall success rate remained low.

% From the above two case studies, we observe that the agent fails in this task primarily because it struggles to navigate the environment effectively due to its lack of knowledge about the target's specific location.
% Even when the agent believes it has identified the plate, it fails to grasp it, revealing issues with its understanding of object locations relative to itself.

\section{Experimental setup details}
\label{sec:details of experimental setup}

In this appendix, we provide more low-level details on the implementation of \projname\ experiments.

\subsection{Multiple image input in LLaVA}
\label{appendix: multi-LLaVA}

The LLaVA architecture itself is compatible with using multiple images as input, but the released model weights do not have the ability to handle multiple images. In other words, when you input multiple images, it will only focus on the contents of the first image. Therefore, we fine-tune LLaVA through multi-image captioning data to enable it to understand multiple images.

We utilized LLaVA's single-image perception capabilities to enhance its multi-image perception. Specifically, we sampled images collected from the environments and employed the pre-trained LLaVA to describe the content of each image. We organized the responses into multi-image captioning data, which we then used to fine-tune LLaVA. After our verification, the fine-tuned LLaVA demonstrated the ability to perceive multiple input images. After fine-tuning with multiple image datasets, LLaVA serves as the basez model for our subsequent retriever training.

\subsection{Parameter settings}

The specific parameter settings in the experiment are shown in Table~\ref{tab:hyperparameters}.

\begin{table}[ht]
\centering
\caption{Hyperparameters of LLaVA fine-tuned by LoRA}
\begin{tabular}{ll}
\toprule
\textbf{Hyperparameters} & \textbf{Value} \\ \midrule
LLaVA\_version & llava-v1.6-mistral-7b \\ 
train\_batch\_size & 32 \\ 
eval\_batch\_size & 8 \\ 
gradient\_accumulation\_steps & 8 \\ 
learning\_rate\_AI2THOR & 2e-5 \\ 
mm\_projector\_lr\_AI2THOR & 2e-5 \\ 
learning\_rate\_LEGENT & 3e-6 \\ 
mm\_projector\_lr\_LEGENT & 3e-6 \\ 
lora\_r & 16 \\ 
lora\_alpha & 32 \\ 
warmup\_ratio & 0.05 \\ 
model\_max\_length & 32768 \\ 
lr\_scheduler\_type & cosine \\ 
vision\_tower & clip-vit-large-patch14-336 \\ \bottomrule 
\end{tabular}
\label{tab:hyperparameters}
\end{table}

\subsection{Details of LLava-Plain}
\label{llava_plain}

Building on existing work~\citep{asaiself}~\citep{sun2023chatgpt} that employs LLM for text retrieval, we use the generation probability of a special token to represent the score for LLaVA-Plain.
In detail, the effectiveness score $s_i$ is measured by the probability of LLaVA-Plain to generate the special token `Yes' and `No', as in Equation~\ref{eq:llava-plain}, where $p\left( \mathrm{Yes/No} \right)$ denoted the probability of LLaVA-Plain to generate Yes or No.
\ref{sec:agent_prompt}.

\begin{equation}
s_i={\frac{p\left( \mathrm{Yes} \right)}{p\left( \mathrm{Yes} \right) +p\left( \mathrm{No} \right) }}
\label{eq:llava-plain}
\end{equation}

In detail, the prompts we use for LLaVA-Plain and \projname\ are presented in prompt 1 of Appendix~\ref{sec:agent_prompt}. 

\section{\aithor\ environment specifics}

\subsection{Setting differences}

Two popular benchmarks, ALFRED~\citep{Shridhar_2020_CVPR}, and ALFWorld~\citep{ALFWorld20} -- derived from  ALFRED -- are both built on \aithor. However, none of them are directly suitable for benchmarking MLLM agents performing real-world tasks.

ALFRED is a multimodal benchmark in \aithor\ that uses fine-grained navigation actions. However, it requires pixel-level masks to specify objects for interaction actions. MLLM lacks the capability to generate such pixel-level masks, making ALFRED incompatible with MLLM agents without adaptation at either side.

ALFWorld adapts ALFRED for LLM agents (\ie text-only) by simplifying it. Firstly, it provides text feedback as observation, detailing objects in the agent's field of view along with their corresponding IDs. Secondly, it simplifies the action space by replacing all navigation actions with the teleportation action `go to' and composite high-level actions like `heat', `clean', and `cool', each involving multiple atomic interactions. For example, the ``cool object a" action is equivalent to: `open the refrigerator', `put object A inside the refrigerator', `close the refrigerator', `open the refrigerator', and `pick up the object A'. These modifications significantly reduce task difficulty, while allowing LLM agents to perform in ALFWorld.

To better evaluate the fine-grained control abilities of MLLM agents in real-world tasks and longer-horizon more realistic tasks, we reject ALFWorld's setting approach, which uses teleportation and composite high-level actions, opting instead for fine-grained navigation actions and finer-grained actions. However, unlike in ALFRED, since MLLMs cannot generate pixel-level masks by default, we allow interaction actions to reference objects using a numerical ID (\eg cup 1) provided by the environment's feedback, instead of a pixel-level mask.

\subsection{Task Decomposition}
\label{appendix_task_decomposition}
% We follow the method as ALFRed to decompose the whole task into multiple sub-tasks. 
% Firstly, we encode the agent and object states, as well as high-level environment dynamics, into Planning Domain Definition Language (PDDL) rules~\citep{aeronautiques1998pddl}.
% We then define task-specific PDDL goal conditions, for example that a heated potato is resting on a table top. Note that the planner encodes the environment as fully observable and has perfect knowledge about world dynamics. 
% Each task is decomposed into several sub-tasks, and in ALFRed, they provide each subtask a instruction by human labelor.

We adopt the ALFRed~\citep{Shridhar_2020_CVPR} method to decompose the entire task into multiple sub-tasks.
In ALFRed, tasks are decomposed as follows:
First, they encode agent and object states, along with high-level environment dynamics, into Planning Domain Definition Language (PDDL) rules~\citep{aeronautiques1998pddl}.
Next, they define task-specific PDDL goal conditions, such as a heated potato resting on a tabletop.
The planner assumes a fully observable environment with perfect knowledge of world dynamics.
Consequently, each task is decomposed into several sub-tasks, with instructions provided for each subtask by human labelers.

We follow the ALFRed method but adjust the decomposition results.
PDDL-based decomposition can result in inconsistent sub-task difficulty.
For example, a sub-task like `pick up object A' or `put object A on object B', which often follows a navigation sub-task.
If the previous navigation sub-task is executed successfully, the current sub-task can be completed in one step according to the PDDL decomposition.
We adjusted the task decomposition by merging sub-tasks that can be completed in one step with adjacent sub-tasks to balance their difficulty.
We will release the benchmark including our tasks and modified sub-tasks.

\subsection{Success Detection}
\label{appendix_success_detection}

In \aithor, we provide the agent with the environment's metadata and feedback in the Success Detection module. After each step is executed, the agent then determines whether the current sub-task is completed using a few-shot approach. For specific prompts, please refer to Appendix \ref{sec:agent_prompt}.

\subsection{Hierarchy examples}

\begin{figure}[ht]
  \centering
  \begin{subfigure}{0.24\textwidth}
    \centering
    \includegraphics[width=\linewidth]{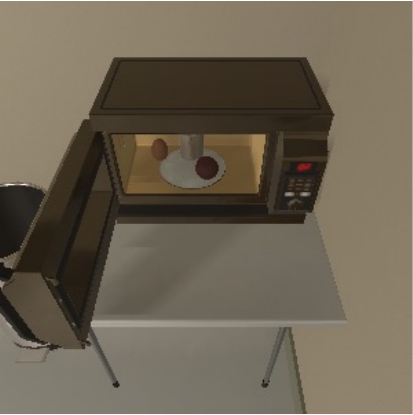}
    \caption{Microwave.} 
    \label{subfig:Microwave}
  \end{subfigure}
  \begin{subfigure}{0.24\textwidth}
    \centering
    \includegraphics[width=\linewidth]{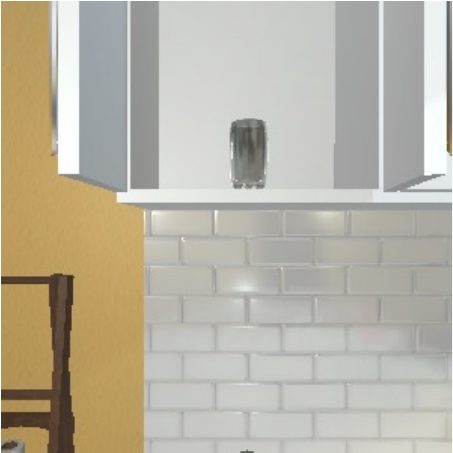}
    \caption{Cabinet.} 
    \label{subfig:Cabinet}
  \end{subfigure}
  \begin{subfigure}{0.24\textwidth}
    \centering
    \includegraphics[width=\linewidth]{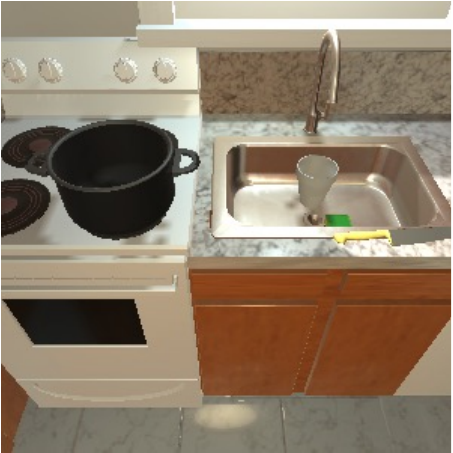}
    \caption{Sink.} 
    \label{subfig:Sink}
  \end{subfigure}
  \begin{subfigure}{0.24\textwidth}
    \centering
    \includegraphics[width=\linewidth]{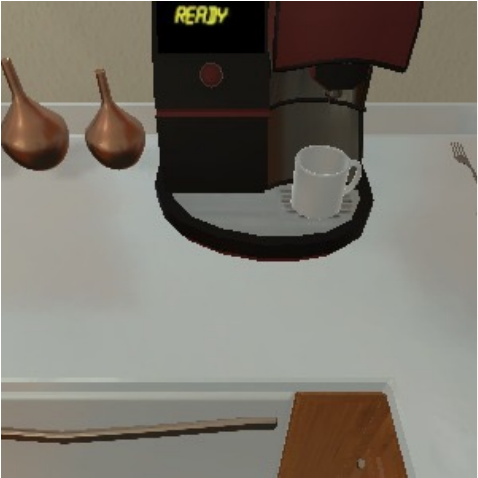}
    \caption{Coffee machine.} 
    \label{subfig:Coffee}
  \end{subfigure}

  \caption{Image examples of object hierarchy in \aithor.} 
  \label{fig:hierarchy}
\end{figure}

Figure~\ref{fig:hierarchy} illustrates multiple cases of the object hierarchy relationship \emph{Inside}, only available in the \aithor\ environment.

\begin{figure}[ht]
  \centering
  \begin{subfigure}{0.24\textwidth}
    \centering
    \includegraphics[width=\linewidth]{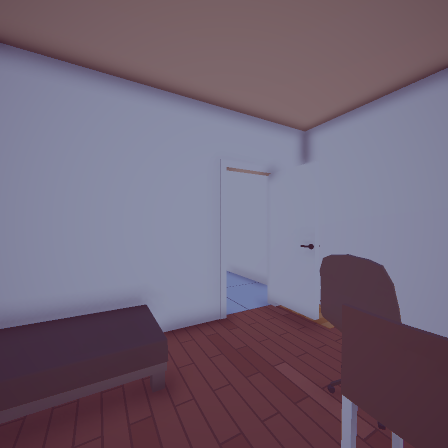}
    \caption{Where-2room.} 
    \label{subfig:Where-2room}
  \end{subfigure}
  \begin{subfigure}{0.24\textwidth}
    \centering
    \includegraphics[width=\linewidth]{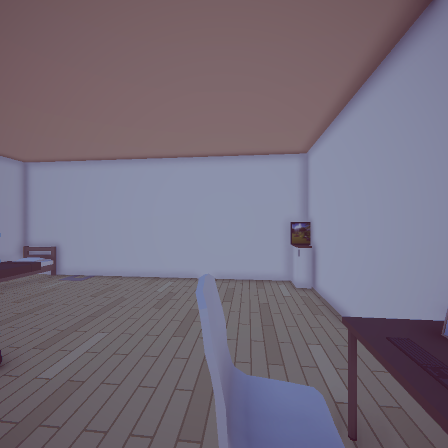}
    \caption{Where-1room.} 
    \label{subfig:Where-1room}
  \end{subfigure}
  \begin{subfigure}{0.24\textwidth}
    \centering
    \includegraphics[width=\linewidth]{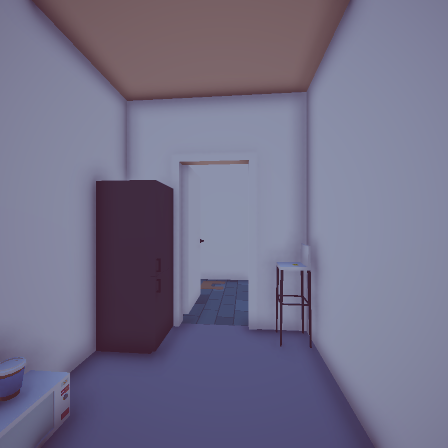}
    \caption{Come-2room.} 
    \label{subfig:Come-2room}
  \end{subfigure}
  \begin{subfigure}{0.24\textwidth}
    \centering
    \includegraphics[width=\linewidth]{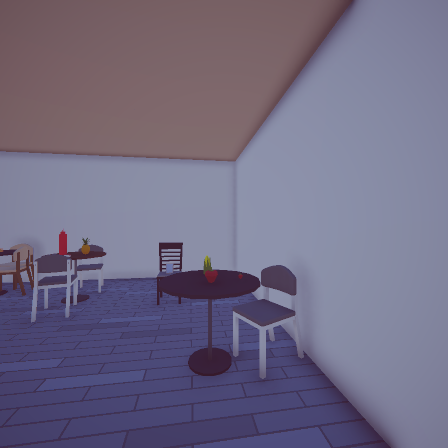}
    \caption{Come-1room.} 
    \label{subfig:Come-1room}
  \end{subfigure}

  \caption{Image examples of different types of tasks in \legent.} 
  \label{fig:legent_subroom}
\end{figure}

Moreover, Figure \ref{fig:legent_subroom} illustrates the different types of rooms in our \legent\ experiments, showing connectivity, but simpler \emph{On} hierarchies than \aithor.

\section{Full step count results}

% Due to space limitations, we add the full tables with average step count results here.
Due to space limitations, we present the full tables with Success Rate (SR) and Average Steps (AS) results here.

Table~\ref{tab:legent_results_complete} shows the performance comparison of different types in LEGENT, with full step count results. While, Table~\ref{tab:combined_ablation_results_complete} shows the performance comparisons of the ablation studies in both AI2-THOR and LEGENT environments, with full step count results.

\begin{table}[ht]
\centering
\caption{Performance comparison of different types in \legent, with full step count results.}
\begin{tabular}{llrrrrr}
\toprule
\textbf{Metric} & \textbf{Type} & \PlainAgent & \textbf{LP} & \SimilarityLLaVA & \textbf{RAP} & \textbf{MART} \\
\midrule
\multirow{5}{*}{\textbf{SR} $\uparrow$} & Where-2room & 0.65   & 0.60   & 0.60   & 0.73  & \textbf{0.88} \\
& Where-1room & 0.78   & 0.80  & 0.80   & 0.89  & \textbf{0.91} \\
& Come-2room  & 0.63   & 0.55  & 0.63    & 0.46  & \textbf{0.73} \\
& Come-1room  & 0.75   & 0.82  & 0.96    & 0.93   & \textbf{0.98} \\
& \textbf{Average} & 0.70   & 0.69   & 0.75   & 0.75   & \textbf{0.87} \\
\midrule
\multirow{5}{*}{\textbf{AS} $\downarrow$} & Where-2room & 25.23  & 28.98  & 28.30 & 26.23 & \textbf{14.03} \\
& Where-1room & 10.40  & 16.49 & 15.36  & 9.82  & \textbf{5.07} \\
& Come-2room  & 32.05  & 34.95 & 28.80  & 35.33 & \textbf{26.90} \\
& Come-1room  & 26.80  & 19.62 & 11.22  & 11.09 & \textbf{9.24}\\
& \textbf{Average} & 23.62  & 25.01 & 20.92 & 20.62 & \textbf{13.81} \\
\bottomrule
\label{tab:legent_results_complete}
\end{tabular}
\end{table}

\begin{table}[ht]
    \centering
    \caption{Ablation studies of \projname\ in the \aithor\ and \legent\ environments, with full step count results.}
    \label{tab:combined_ablation_results_complete}
    \begin{tabular}{lllrrr}
    \toprule
    \textbf{Environment} & \multicolumn{2}{c}{\textbf{Metric}} & \textbf{w/o Abstraction} & \textbf{Sim.+FTM} & \textbf{MART} \\
    \midrule
    \multirow{4}{*}{\textbf{\aithor}} & \multicolumn{2}{c}{\textbf{SR} $\uparrow$} & 0.31 & 0.34 & \textbf{0.40} \\
     & \multicolumn{2}{c}{\textbf{SR-Sub} $\uparrow$} & 0.73 & 0.74 & \textbf{0.75} \\
     & \multicolumn{2}{c}{\textbf{AS} $\downarrow$} & 81.22 & 78.09 & \textbf{78.48} \\
     & \multicolumn{2}{c}{\textbf{AS-Sub} $\downarrow$} & 22.47 & 21.60 & \textbf{21.71} \\
     \midrule

     \multirow{10}{*}{\textbf{\legent}} &  \multirow{5}{*}{\textbf{SR} $\uparrow$} & Where-2room & 0.78   & 0.72   & \textbf{0.88} \\
     & & Where-1room & 0.74   & 0.89 & \textbf{0.91} \\
     & & Come-2room   & 0.63  & 0.55 & \textbf{0.73} \\
     & & Come-1room  & 0.95   & 0.91 & \textbf{0.98} \\ 
     & & \textbf{Average} & 0.77 & 0.77 & \textbf{0.87} \\ \cline{2-6}

      &  \multirow{5}{*}{\textbf{AS} $\downarrow$} & Where-2room & 17.36 & 15.96 & 14.03 \\
     & & Where-1room & 12.36 & 11.51 & 5.07  \\
     & & Come-2room  & 32.75 & 34.13 & 26.90 \\
     & & Come-1room  & 11.47 & 13.73 & 9.24 \\
     & & \textbf{Average} & 18.48 & 18.83 & 13.81 \\
    \bottomrule
    \end{tabular}
\end{table}

% \section{Hyper-parameter Details}
\section{Algorithm}

\projname's Agent Execution Pseudocode is shown in Algorithm \ref{algorithm:main}.

\begin{CJK*}{UTF8}{gkai}

\begin{algorithm}[H]
    \caption{\projname\ Agent Execution Pseudocode}
    \label{algorithm:main}
    \begin{algorithmic}[1] % 每行显示行号
    \Require Expert Trajectory Memory $\mathcal{M}$, Retriever $q_\theta$, Policy $\pi$, Task $\ell^c$, Horizon $H$, Initial Observation $o_1^c$, Preference Pairs $\mathcal{D}$
    \Ensure the Success status of task execution
    
    \State Fine-tune retriever $q_\theta$ with Preference Pairs$\mathcal{D}$
    \State Retrieve reference trajectory $\tau^e \leftarrow q_\theta(\ell^c, o_1^c, \mathcal{M})$
    \State \textit{TrajectoryAbstraction} to simplify $\tau^e$

    \For{$t = 1$ to $H$}
    
        \If{$t != 1$}
            \State $r_t \leftarrow $ \textit{SelfReflection}$(a_{t-1}, o_{t-1}^c, f_{t-1}^c)$ 
            \State Select action $a_t \leftarrow \pi(\ell^c, \tau^e, o_t^c, r_t)$
        \EndIf
        \If{$t == 1$}
            \State Select action $a_t \leftarrow \pi(\ell^c, \tau^e, o_t^c)$
        \EndIf
        
        \State $(o_{t}^c, f_{t}^c) \leftarrow \textit{ActionExecution}(a_t)$
        \If{\textit{SuccessDetection}$(\ell^c, f_{t+1}^c)$}
            \State \Return \textbf{True} \hspace{0.5em} // Task successfully completed
        \ElsIf{$t \ge H$}
            \State \Return \textbf{False} \hspace{0.5em} // Task failed after reaching horizon
        \EndIf
    \EndFor
    \end{algorithmic}
\end{algorithm}
\end{CJK*}

%\begin{figure}[t]
%    \centering
%    % \vspace{-4mm}
%    \includegraphics[width=.9\linewidth]{ICLR 2025 Template/images/comparison.pdf}
%    \caption{Similarity-Based Retriever vs. \projname. Traditional retrieval methods (1) depend on calculating weighted sums of image and text embedding similarities. In contrast, our approach (2) introduces interactive learning to assess the relevance between the current and expert trajectories.}
%    \label{fig:comparison}
%    % \vspace{-32pt}
%\end{figure}

\clearpage
\section{Agent prompts} \label{sec:agent_prompt}

\begin{lstlisting}[breaklines=true,caption={Retriver Prompt to LLaVA.}]
You are a highly intelligent vision language assistant agent situated in a virtual environment. 
You have been given a task instruction that you need to complete. 
Additionally, you are provided with a reference trajectory, which includes previous task instructions, actions, and egocentric observations from the same virtual environment. 
This reference trajectory represents a successful completion of a task and is intended to guide you in performing the current task.
The current task instruction is: [current task].
The task instruction of reference trajectory is [memory task].
Among these input images, the 1st image is your current observation, while the other images are milestones of observations from the reference trajectory. 
The abstraction of the reference trajectory is: 

<Description of Milestone 0>, <Image 0>, <Feedback 0>, <Action 0>;
<Description of Milestone 1>, <Image 1>, <Feedback 1>, <Action 1>;
...


Your should thoroughly understand the current task and the reference trajectory. Then, analyze whether the reference trajectory can assist in executing the current task by answering 'Yes' or 'No'.
You should only respond in the format described below, and you should not output comments or other information: 
Answer: Yes or No.
\end{lstlisting}
\clearpage

\begin{lstlisting}[breaklines=true,caption={Trajectory Abstraction Prompt.}]
You are a highly intelligent vision-language assistant agent placed within a virtual environment. You are provided with a trajectory consisting of:
- Trajectory Task Instruction: {Task Instruction}
- A sequence of first-person perspective observations (<Image x>)
- A sequence of environment feedbacks (<Feedback x>)
- A sequence of actions (<Action x>)
- Another Task Instruction: {Current Task Instruction}

Your Tasks:
1. Fully comprehend the tasks accomplished in the trajectory.
2. Identify the significant milestones in the trajectory that are essential for accomplishing the task. These are points where important decisions are made, goals are achieved, or notable changes in the environment or state occur. Do not treat every image as a significant milestone. If the target object of Another Task Instruction appears in the trajectory feedback, then the point where it appears is also a significant milestone because it is very important to Another Task Instruction.
3. For each significant milestone, provide:
- A description of the milestone.
- The corresponding image (<Image x>).
- The corresponding feedback (<Feedback x>).
- The sequence of actions taken between this milestone and the next one.

<Image 0>, <Feedback 0>, <Action 0>;
<Image 1>, <Feedback 1>, <Action 1>;
...

Response Format (do not include any comments or additional information):
1. {Description of significant milestone 1}: <Image a>. <Feedback a>. Actions: {Actions taken between this significant milestone and the next significant milestone} (such as <Action a>, <Action b>).
2. {Description of significant milestone 2}: <Image c>. <Feedback c>. Actions: {Actions taken between this significant milestone and next significant milestone} 
...

Notes:
- Ensure that the number of <Image x> and <Action x> matches the provided trajectory.
- Only include the specified information in your response.
- A significant milestone is a point in the trajectory where a key part of the task is accomplished. If the target object of Another Task Instruction appears in the trajectory feedback, then the point where it appears is also a significant milestone because it is very important to complete Another Task .
\end{lstlisting}
\clearpage

\begin{lstlisting}[breaklines=true,caption={Self-Reflection Prompt.}]
You are a vision language assistant agent with high intelligence.
You are placed inside a virtual environment, equipped to handle a wide range of tasks in the virtual environment.
Your advanced capabilities enable you to process and interpret egocentric observation screenshots and environment feedback.
Your task is to examine these inputs, interpret the environmental feedback, and determine whether the executed action takes effect.

Current task:
{task_instruction}

Previous actions:
{previous_action_1}, {previous_action_2}, ...

Previous environment feedback:
{previous_feedback}

Reasoning for the previous actions:
{previous_reasoning_1}, {previous_reasoning_2}, ...

Previous observations:
{previous_image_1}, {previous_image_2}, ...

Reasoning: You need to answer the following questions step by step to get some reasoning based on the previous actions and sequential images of the execution of the previous actions.
1. What is the last executed action?
2. Was the last executed action successful? Give reasons. You should refer to the following rules:
- If the action involves movement of position, change of view, or interaction with an object, it is considered unsuccessful when the image you currently observe remains unchanged as the previous frame.
3. If the last action is not executed successfully, what is the most probable cause? You should give only one cause and refer to the following rules:
- If it is an interaction action, the most probable cause was that the object id of the interaction was wrong.
- If it is a movement action, the most probable cause was that you were blocked by seen or unseen obstacles.
4. If the last action is executed successfully, Does the previous action sequence promote the progress of the current task?
5. If the answer to Reasoning Question 4 is No, how should you adjust to promote the progress of the current task?

You should only respond in the format described below, and you should not output comments or other information:
Reasoning:
1. ...
2. ...
3. ...
4. ...
5. ...
...
\end{lstlisting}
\clearpage

\begin{lstlisting}[breaklines=true,caption={Action Planning Prompt.}]
You are a robot working in a household environment. You can move and interact with the objects you see.
The actions you can perform include: 
1. 'MoveAhead': Move one step forward.
2. 'RotateLeft_$degree': Turn to the left by the specified number of degrees, ranging from 0 to 180 degrees.
3. 'RotateRight_$degree': Turn to the right by the specified number of degrees, ranging from 0 to 180 degrees.
4. 'LookUp': Look up 30 degrees.
5. 'LookDown': Look down 30 degrees.
6. 'Take $objectID': You can take any object in your line of sight. The $objectID can be obtained from the environment feedback.
7. 'Put $objectID on/in $targetID': You can put the object in your hand onto/into the target receptacle. The $targetID can be obtained from the environment feedback, and $objectID refers to the object in your hand.
8. 'Open $objectID': You can open any openable object in your line of sight. The $objectID can be obtained from the environment feedback.
9. 'Close $objectID': you can close any open object in your line of sight. The $objectID can be obtained from the environment feedback.
10. 'ToggleOn $objectID': You can toggle on the switch of the object, such as a faucet or microwave. The $objectID can be obtained from the environment feedback.
11. 'ToggleOff $objectID': You can toggle off the switch of the object, such as faucet or microwave. The $objectID can be obtained from the environment feedback.
12. 'Slice $objectID': You can slice any object in your line of sight. The $objectID can be obtained from the environment feedback.

Examples of actions: RotateLeft_30; MoveAhead; Take mug 1; Open fridge 1; ToggleOn microwave 1; Close fridge 1;...

You need to follow the task instructions to complete the task. Here is some helpful information.

Current task:
{task}

Current environment feedback:
{obs}

Previous environment feedback:
{previous_obs}

Previous action and reasoning:
{previous_action}

Current observation
<image>

Previous observation
<image>

Reference trajectory abstraction: The reference trajectory is a successful trajectory, which is used to guide you to complete the current task. The task of reference trajectory is {current task}. 

<Description of Milestone 0>, <Image 0>, <Feedback 0>, <Action 0>;
<Description of Milestone 1>, <Image 1>, <Feedback 1>, <Action 1>;
...

Based on the above information, you should first analyze the current situation, and provide the reasoning for what you should do for the next step to complete the task. Then, you should output the exact action you want to execute in the simulator.

Reasoning: You should think step by step and provide detailed reasoning to determine the next action executed on the current state of the task. You need to answer the following questions step by step.

1. Does reference trajectory abstraction exist? If the answer is no, ignore the questions from number 2 to number 4.
2. What process does reference trajectory abstraction describe?
3. Consider what is your current task. Based on the Observation of the previous step and Current Observation, which waypoint has the current task reached?
4. Based on the answer of the question number 3, you should consider how the current waypoint and the parts after that in the reference trajectory abstraction can help you with your current task. The help provided by the reference trajectory abstraction can be knowing the location of the target object or knowing the execution flow of a combined action.
5. Based on the completion progress of the current task and the answer to question number 4, what should you do for the next step?
6. Why do you take this action next step?

Action: The best action to execute next to progress in completing the task. You should pay more attention to the following action rules:
1. Given the current situation and task, you should only choose the most suitable action from the valid action set. You cannot use actions that are not in the valid action set to control the application, especially 'Await Next Task'.
2. If the Action of the previous step fails, you should not continue trying but should consider adjusting your position to get closer to the target object.
3. You MUST NOT match or imitate the reference trajectory. You should think about how to complete the current task based on the answer to Reasoning Question 4.

You should only respond in the format described below, and you should not output comments or other information:
Reasoning:
1. ...
2. ...
3. ...
4. ...
5. ...
6. ...

Action: ...

\end{lstlisting}
\clearpage

\begin{lstlisting}[breaklines=true,caption={Success Detection Prompt.}]
You are a highly intelligent vision-language assistant agent.
You are situated in a virtual environment, equipped to handle a diverse array of tasks.
Your advanced capabilities allow you to process and interpret egocentric observation screenshots, environmental feedback and environmental metadata.
Your task is to examine these inputs, understand the environmental metadata, and assess the success of the current task.

Current task:
<task>

Environmental Metadata
<environment metadata>

Environmental Feedback
<environment feedback>

Current Inventory
<inventory>

You need to refer to the following rules:

1. If the current task contains multiple tasks, it is considered successful only when each task succeeds.
2. If the task is a navigation task, you need to check the environmental metadata. The navigation task succeeds when the target object is in view and the distance is less than 1m.
3. If the task is a pickup task, then according to the environmental feedback, the pickup task succeeds when the target object is in your inventory.
4. If the task is a put down task (put object a on/in object b), the put down task succeeds when the environmental feedback from the environment includes You put object a on/in the object b successfully.
5. If the task is a clean task, the clean task succeeds when the cleaned_objects in environmental metadata include the target object and the same target object is also in inventory.
6. If the task is a cool task, the cool task succeeds when the cooled_objects in environmental metadata includes the target object and the same target object is also in inventory.
7. If the task is a heat task, the heat task succeeds when the heated_objects in environmental metadata includes the target object and the same target object is also in inventory.

You should only respond in the format described below, and you should not
output comments or other information:

Answer: True or False.
\end{lstlisting}

% \section{Diagram Legends} \label{sec:data_format}

% \begin{compactitem}
% % 这里我们详细
%     \item Current Observation (\includegraphics[width=0.3cm]{ICLR 2025 Template/images/icons/observation.pdf}): Current Image + Current Feedback. Feedback contains the items present in the field of view and the items in the hand.
%     \item Current Task (\includegraphics[width=0.4cm]{ICLR 2025 Template/images/icons/task.pdf}): Task Instruction + Initial Observation.
%     \item One Trajectory (\includegraphics[angle=90,width=0.6cm]{ICLR 2025 Template/images/icons/trajectory.pdf}): Trajectory Instruction + Image Sequence + Feedback Sequence + Action Sequence.
%     \item One Abstract Trajectory (\includegraphics[angle=90,width=0.6cm]{ICLR 2025 Template/images/icons/abstract_trajectory.pdf}): Trajectory Instruction  + Milestone Description + Abstract Image Sequence + Abstract Feedback Sequence + Overarching Action Sequence.
% \end{compactitem}

\end{document}